\documentclass[sigconf,nonacm]{acmart}

\usepackage{multirow}
\usepackage{microtype}
\usepackage{graphicx}
\usepackage{subcaption}
\usepackage{booktabs} 
\usepackage{array}
\usepackage{graphicx}
\usepackage[dvipsnames]{xcolor}
\usepackage{color-edits}
\usepackage{enumitem}

\usepackage{algorithm}
\usepackage{algorithmic}
\addauthor[Dominik]{dr}{orange}
\addauthor[Elisabeth]{ep}{blue}
\addauthor[Kirk]{kb}{ForestGreen}
\addauthor[Gauri]{gj}{purple}
\setcopyright{acmlicensed}
\copyrightyear{2026}
\acmDOI{XXXXXXX.XXXXXXX}
\begin{document}
\settopmatter{printfolios=true}
\title{CTRL Your Shift: Clustered Transfer Residual Learning for Many Small Datasets}

\author{Gauri Jain}
\affiliation{%
  \institution{Harvard University}
  \city{Cambridge}
  \state{Massachusetts}
  \country{USA}
}
\author{Dominik Rothenhäusler}
\affiliation{%
  \institution{Stanford University}
  \city{Stanford}
  \state{California}
  \country{USA}
}
\author{Kirk Bansak}
\affiliation{%
  \institution{University of California, Berkeley}
  \city{Berkeley}
  \state{California}
  \country{USA}
}
\author{Elisabeth Paulson}
\affiliation{%
  \institution{Harvard University}
  \city{Cambridge}
  \state{Massachusetts}
  \country{USA}
}








\begin{abstract}
  Machine learning (ML) tasks often utilize large-scale data that is drawn from several distinct sources, such as different locations, treatment arms, or groups. In such settings, practitioners often desire predictions that not only exhibit good overall accuracy, but also remain reliable within each source and preserve the differences that matter across sources. For instance, several asylum and refugee resettlement programs now use ML-based employment predictions to guide where newly arriving families are placed within a host country, which requires generating informative and differentiated predictions for many and often small source locations. However, this task is made challenging by several common characteristics of the data in these settings: the presence of numerous distinct data sources, distributional shifts between them, and substantial variation in sample sizes across sources. This paper introduces Clustered Transfer Residual Learning (CTRL), a meta-learning method that combines the strengths of cross-domain residual learning and adaptive pooling/clustering in order to simultaneously improve overall accuracy and preserve source-level heterogeneity. We establish new theory showing that high-quality clusters can be learned efficiently, bypassing the need for repeated model refitting over candidate subsets. We evaluate CTRL alongside other state-of-the-art benchmarks on 5 large-scale datasets. This includes a dataset from the national asylum program in Switzerland, where the algorithmic geographic assignment of asylum seekers is currently being piloted. CTRL consistently outperforms the benchmarks across several key metrics and when using a range of different base learners.
\end{abstract}





\maketitle

\section{Introduction}
Many machine learning (ML) tasks draw data from several distinct sources.  A source can be a physical site, a time period, a treatment arm, or any other index that partitions the data. Sources may appear as a categorical feature inside one large data set (e.g. hospital ID, month of year, medical treatment arm), or each source may contribute its own smaller data set collected under local conditions. In any of these settings, practitioners may desire more than predictive accuracy from an ML model: predictions should remain reliable within every source and preserve the differences that matter across sources, so that predictions remain credible for downstream ranking, assignment, and recommendation tasks. 

As a concrete example, several asylum and refugee resettlement programs use ML‑based employment predictions to guide where newly arriving families are placed within a host country \citep{Bansak_2024,Ahani_2021}. In this setting, the relevant sources are \emph{locations}, and we will use that terminology going forward. Each candidate location (e.g. city, region, state) differs in labor‑market conditions, support infrastructure, and demographics. These factors, in combination with the refugees' personal characteristics, in turn influence refugees' employment outcomes. Accurate and \emph{location‑specific} predictions therefore matter for generating high quality assignment policies.

Building models that have high predictive accuracy \emph{and} differentiate between locations is difficult for two reasons. First, locations can drastically vary in size. In resettlement programs, geographic quotas (or historical arrival patterns) allocate families unevenly across locations, producing training datasets that range anywhere from 50-4000 rows. Small locations inevitably suffer higher estimation error. Second, covariate and outcome distributions differ across locations. Thus, naively pooling all data (henceforth, \emph{global models}) can blur these distribution shifts, while training separate models (henceforth, \emph{local models}) ignores valuable shared structure.

These problem features and challenges hint at two possible adaptive meta-learning strategies that have become increasingly prominent in recent years. However, there are some notable potential tradeoffs between the two strategies. First, a number of procedures proposed in recent research leverage cross-domain residual learning as a way of tailoring ML models towards specific target data sets or data sources \citep[e.g.][]{long2016unsupervised,khan2019cross,K_nzel_2019}. This aligns with our prioritization of informative location-specific predictions. However, a common weakness of residual learning procedures is that they can often be unreliable for target data sets that are too small. In settings such as ours, the possibility of many small locations potentially limits the effectiveness of residual learning.

Second, other recent research has highlighted the value of adaptive pooling/clustering for handling a collection of data sets or data sources \cite[e.g.][]{pmlr-v252-shen24a,jeong2024out,huang2025stability}. The advantage offered by this strategy is in borrowing strength across locations, bolstering predictive performance even for small locations. When pooling is performed with \emph{overall} accuracy in mind, however, such an approach may fail to capture heterogeneity specific to different locations. Accounting for this type of heterogeneity is vital in a setting such as ours, where differentiated predictions across locations are foundational for guiding downstream tasks. 

\begin{figure}[ht!]
  \centering
\includegraphics[width=1\linewidth]{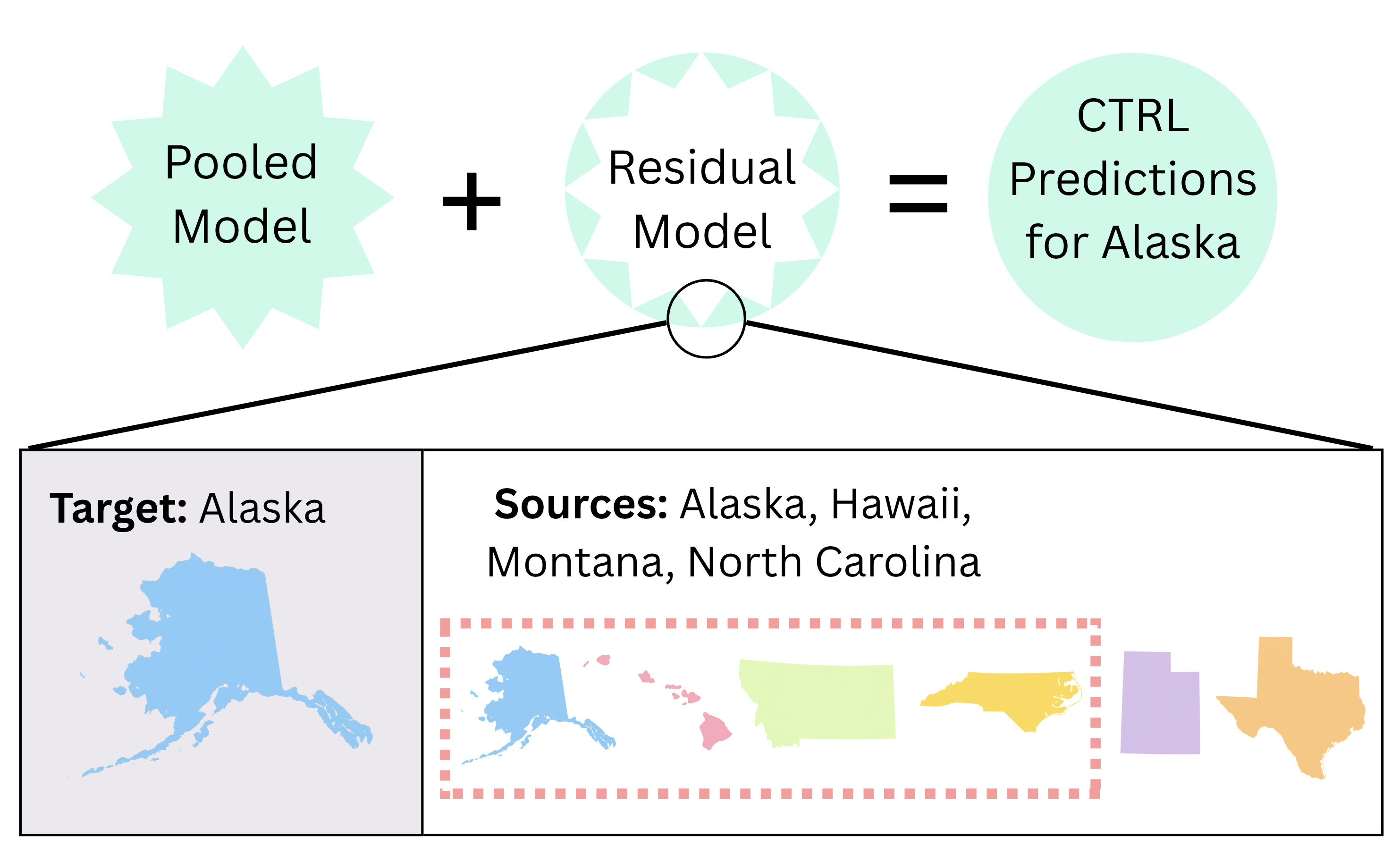}
  \caption{An overview of CTRL on the example task of  predicting educational attainment in the state of Alaska. CTRL's predictions are the sum of (1) predictions from a pooled model using all education data from the US and (2) predictions from a residual model that clusters location datasets from Alaska, Hawaii, Montana, and North Carolina.}
  \label{fig:clustering_visual}
  \vspace{-0.4cm}
\end{figure}

In this paper, we propose a meta-learning approach that seeks to leverage the strengths of both strategies by navigating their tradeoffs in a principled fashion. First, we anchor our approach by establishing a baseline procedure---which we refer to as \emph{transfer residual learning (TRL)}---that trains a global model on the pooled dataset, and then fine-tunes location-specific residual models to capture systematic deviations. This two-stage baseline approach leverages the pooled dataset for generalization while allowing location-level corrections. However, performance is limited by location size: small locations may still suffer from high variance in their fine-tuning step. To address this limitation, we propose \emph{Clustered Transfer Residual Learning (CTRL)}, a new meta-learning algorithm that identifies clusters of locations that improve a target location's fine-tuning step (See Figure \ref{fig:clustering_visual}). By adaptively clustering locations, CTRL automatically adapts to location similarity and data availability. When no meaningful cluster exists for a location, the method defaults to baseline transfer residual learning. This adaptive design enables CTRL to achieve more accurate predictions, particularly for small locations, while retaining predictive differentiation. Furthermore, CTRL is agnostic to the base learner used for model building (e.g. linear regression, random forest, etc.).

    We evaluate CTRL, TRL, and additional state-of-the-art benchmarks on 5 large-scale datasets---including a dataset from the national asylum program in Switzerland, where the algorithmic geographic assignment of asylum seekers is currently being piloted \citep{Bansak_2024}. For three datasets, we evaluate three metrics: (1) rank-weighted average (RWA), (2) overall mean squared error (MSE), and (3) MSE among small locations. Inspired by \citet{Yadlowsky02012025}, RWA captures how well each location can identify top-performing individuals, serving as a proxy for a model's ability to perform well on downstream tasks such as ranking, assignment, and allocation. Hence, RWA is the most directly relevant metric in a policy learning context.
For the remaining two datasets, we report only the MSE metrics, as the contexts in those datasets do not support a meaningful notion of ranking.  
Our results show that CTRL consistently outperforms the benchmarks in all three metrics. This strong performance demonstrates the value of considering CTRL for deployment in algorithmic asylum and refugee resettlement efforts, among other possible settings.

\section{Contributions}
We introduce \emph{Clustered Transfer Residual Learning (CTRL)}, a meta-learning algorithm for the many-locations prediction problem that addresses both distribution shift and data scarcity. CTRL clusters sources based on \emph{conditional} similarity—captured through residuals—enabling targeted pooling that adapts to heterogeneity in $P(Y \mid X)$. We provide a theoretical analysis that motivates the clustering objective employed in CTRL and empirically evaluate its performance relative to more generic clustering approaches. Code and datasets are publicly available\footnote{\url{https://github.com/Gjain234/CtrlYourShift}}. We also benchmark CTRL against global, local, and residual-learning baselines as well as two state-of-the-art methods used for multi-source data, across five datasets and multiple model classes (linear, tree-based, and ensemble models). CTRL consistently improves decision-quality metrics and reduces prediction error, outperforming the deployed methods that have been piloted in the Swiss asylum seeker system.

\paragraph{Novelty.}
Our contributions are novel along six axes:
\begin{enumerate}[leftmargin=14pt]
    \item  \textit{Residual-level clustering.} We propose a clustering criterion that groups sources by residual similarity rather than covariate distance or feature embeddings. This model-agnostic approach directly targets the predictive signal and, to our knowledge, is the first to use explicit residual-level clustering to guide selective transfer across sources.
    \item \textit{Theory-informed cluster learning.} We provide new theory showing that high-quality clusters can be learned efficiently, without repeatedly refitting models for every candidate subset. \item \textit{Excess-risk results.} We further provide an out-of-distribution excess-risk bound under distribution shift that characterizes when pooling reduces variance versus when shift-induced error dominates, clarifying when transfer helps or hurts.
    \item \textit{Outperformance over naive clustering.} We compare our clustering approach to Wasserstein distance and correlation baselines and find significantly improved recovery of true clusters, indicating that generic distance measures may not always align with predictive relevance for heterogeneous datasets. 
    \item \textit{Many-sources and ranking evaluation regime.} We study the many-sources regime, where dozens of small, heterogeneous datasets challenge both standard meta-learning and distribution-shift methods, and we evaluate performance using deployment-relevant decision-quality metrics rather than MSE alone. 
    \item \textit{Unified residual transfer and adaptive pooling.} CTRL provides a principled integration of residual transfer and adaptive pooling: the optimization stabilizes small sources, selects cluster size, and naturally reverts to TRL when pooling would induce bias.
\end{enumerate}

\section{Related Work}

\textit{Fine Tuning and Residual Learning.} Our approach leverages a form of model stacking or residual learning, which has a long history in machine learning \cite{wolpert1992stacked}, and is related to popular approaches such as gradient boosting \cite{friedman2001greedy} and fine-tuning \cite{yosinski2014transferable}. The proposed approach combines a form of residual learning with data-set adaptive pooling, to improve performance for small data sets under distribution shift.

\textit{Robust machine learning.} \citet{zhang2024minimaxregretestimationgeneralizing} focus on robustifying measurements of treatment effects by learning heterogeneous treatment predictions that work well across multiple domains. Group distributionally robust optimization (Group DRO) \citep{sagawa2020distributionallyrobustneuralnetworks} learns a prediction model by optimizing the worst-case group risk. \citet{xie_doremi_2023} propose training a small proxy model using Group DRO to select domain weights. In our setting, such approaches would yield the same prediction for every location, thereby undermining the downstream decision task of leveraging location-specific synergies. Simple data-set balancing via re-weighting (RWG) has been shown to have competitive worst-group accuracy \cite{idrissi2022simpledatabalancingachieves}. Just Train Twice (JTT) \cite{liu2021justtraintwiceimproving} is designed to improve worst-case performance. However, such procedures can exhibit overly conservative performance in terms of average-case MSE.

\textit{Causal Inference.} In causal inference, the focus is often on learning the contrast between a treatment and control group, which can be seen as two different distributions.
Thus, many heterogeneous treatment effect estimators can be re-framed as cross-domain residual learning strategies.  \citet{nie2020quasioracleestimationheterogeneoustreatment} provide a loss minimization framework for heterogeneous treatment effect estimation based on Robinson's decomposition. Our approach is  related to \citet{K_nzel_2019}, who combine multiple residual learning models for heterogeneous treatment effect estimation. However, this type of residual learning is not appropriate in the many-data set setting, since it ignores the shared structure across multiple small data sets.   

\textit{Dataset pooling under distribution shift.} Given the breadth of work on data pooling, we restrict our discussion to recent developments. In \citet{pmlr-v252-shen24a}, the authors discuss how aggregating data from many diverse sources can sometimes harm model performance due to distribution shifts—a challenge they term the Data Addition Dilemma. They propose heuristics to guide data selection for improved accuracy and robustness. Our procedure is similar in the sense that we also adaptively add data sets that are ``close" to the target in some sense, but we measure the closeness in terms of residual unexplained variance instead of the joint $(X,Y)$ distribution. \citet{ye_data_2024} use data scaling laws to identify optimal data mixtures; however, the proposed method does not scale to settings with a large number of datasets. Furthermore, \citet{ye_data_2024} address a different performance–speed tradeoff relevant to large language model training, which differs from the setting considered here. 

\textit{Data Generation Techniques.} Oversampling methods such as SMOTE \cite{smote}, data generation techniques like mixup \cite{zhang2018mixupempiricalriskminimization} and interpolation based approaches \citep{fan2023generatingsyntheticdatasetsinterpolating} offer solutions for mitigating class imbalance by artificially augmenting the dataset. However, these approaches are often not viable in real-world deployments due to the potential liabilities associated with generating synthetic data. Additionally, when certain subgroups contain lower-quality or noisier data, such techniques may reduce robustness by amplifying the influence of unreliable samples through oversampling.

\section{Methodology}
\subsection{Preliminaries}
We consider a training dataset $\mathcal{D}_{\text{train}} = \{(X_i, M_i, Y_i)\}_{i=1}^{n_{\text{train}}}$ of size $n_{\text{train}}$, where each individual $i$ is associated with a feature vector $X_i \in \mathcal{X}$, a location designation $M_i\in\mathcal{M}$, and a response variable $Y_i \in \mathcal{Y}$. We assume that location designation forms a partition of the training data into $|\mathcal{M}|$ disjoint groups such that each individual belongs to exactly one location. Let $\mathcal{D}_{\text{train}}^g = \{(X_i, M_i, Y_i) : M_i = g\}$ be the subset of the data corresponding to location $g \in \mathcal{M}$. For this paper, we set $\mathcal{M}$ to represent locations, but $\mathcal{M}$ could more generally denote any (mutually exclusive and collectively exhaustive) set of data sources, groupings, or partitions.

We aim to learn a predictive model $\hat{f}: \mathcal{X} \times \mathcal{M} \to \mathcal{Y}$ from the training data, which maps features and location to predicted outcomes.

In addition to the training data, we consider a test dataset $\mathcal{D}_{\text{test}} = \{(X_i, M_i, Y_i)\}_{i=1}^{n_{\text{test}}}$ consisting of $n_{\text{test}}$ individuals, which we use to evaluate the performance $\hat{f}$.

\subsection{Residual Learning} \label{xlearn}
Residual learning follows a two-stage meta-learning framework, where each component model may be instantiated with any supervised learning method. 
We call this model $\hat{f}_{TRL}$ for Transfer Residual Learning because it estimates residuals using a combination of pooled and location-specific data.

In the first stage, we use the entire training dataset to train a base model $\hat{f}_{\text{base}}: \mathcal{X} \times \mathcal{M} \rightarrow \mathcal{Y}$ to predict the outcome $Y_i$ using covariates $X_i$, and location designation $M_i$: $\hat{f}_{\text{base}}(X_i,M_i) \approx \mathbb{E}[Y_i \mid X_i, M_i]$. 
In the second stage, we refine the base model through residual learning to capture location-specific heterogeneity. For an individual $i$ belonging to location $g$, we define the residual as:
$R^g_i = Y_i - \hat{f}_{\text{base}}(X_i,g)$. 

For each location $g \in \mathcal{M}$, we train a residual model $\hat{f}^{g}_{\text{residual}}: \mathcal{X} \rightarrow \mathbb{R}$ using only individuals with $M_i = g$, aiming to satisfy $\hat{f}^{g}_{\text{residual}}(X_i) \approx R_i^g$ for all $i$ in location $g$. The final prediction for individual $i$ in location $g$ is given by
$$\hat{f}_{\text{TRL}}(X_i, g) = \hat{f}_{\text{base}}(X_i, g) + \hat{f}^{g}_{\text{residual}}(X_i),$$
which combines the global base model with a location-specific correction.

\subsection{Clustered Transfer Residual Learning}\label{sec:cluster-def}
CTRL extends the residual learning framework by replacing location-specific residual models with cluster-specific residual models. Each cluster consists of one or more locations, enabling shared learning across structurally similar or data-sparse subpopulations (see Figure \ref{fig:clustering_visual} as an example). Let $\mathcal{C}(g) \subseteq \mathcal{M}$ denote the cluster constructed for location $g$. We define $\hat{f}_{\text{residual}}^{\mathcal{C}(g)}$ as the residual model trained using all training data from locations in $\mathcal{C}(g)$. The final prediction for an individual $i$ in location $g$ is:
$$\hat{f}_{\text{CTRL}}(X_i, g) = \hat{f}_{\text{base}}(X_i, g) + \hat{f}_{\text{residual}}^{\mathcal{C}(g)}(X_i).$$

Our decision for clustering this way is supported by Proposition~\ref{prop:leafwise-ctrl-risk}, which provides a theoretical justification that similarity should be defined through residual distributions rather than through feature distance or broader distributional metrics. Based on this, we now formalize our approach that clusters on conditional residual similarity, jointly stabilizing small sources, adapting cluster size, and reverting to TRL or the global model when appropriate. This formulation is intuitive, theoretically grounded, model-agnostic, and generalizes across datasets. 

To conceptualize the problem of constructing clusters that we face, let $h^g(r \mid X)$ denote the conditional distribution of residuals $r = Y - \hat{f}_{\text{base}}(X,g)$ for location $g$. If two locations $g_1$ and $g_2$ satisfy $h^{g_1}(r|X=x) = h^{g_2}(r|X=x)$ for all $x$, then pooling should decrease error, since it does not introduce bias. When $h^{g_1}$ and $h^{g_2}$ differ, pooling their data may help or harm the residual model for location $g_1$, depending on the similarity of their distributions and relative sizes. This tradeoff is especially pronounced when one location is much larger. 
In order to properly navigate this tradeoff in a principled empirical manner, we propose a new optimization-based objective. We provide theoretical motivation for this objective in Section \ref{sec:theory}. In addition, we also consider clustering strategies based on generic distributional distances, such as using Wasserstein distance \cite{wass}, and show how they are empirically ineffective compared to our theoretically grounded optimization-based approach (Section \ref{sec:clusteranalysis}).

To construct clusters, we focus on a target location $g \in \mathcal{M}$. For each location, searching over all $2^{|\mathcal{M}|}$ possible clusters is computationally infeasible, so we develop a data-driven heuristic for identifying effective clusters. We partition the training data into a random 80/20 split, denoted $\mathcal{D}_{\text{train}}^{80}$ and $\mathcal{D}_{\text{train}}^{20}$. On $\mathcal{D}_{\text{train}}^{80}$, we train $\hat{f}_{\text{base}}$ and location-specific residual models $\hat{f}^m_{\text{residual}}$ for all locations $m \in \mathcal{M}$. For each individual $i \in \mathcal{D}_{\text{train}}^{g,20} := \{ i \in \mathcal{D}_{\text{train}}^{20} \mid M_i = g \}$, we compute $r_{im} = \hat{f}^m_{\text{residual}}(X_i)$ for all $m \in \mathcal{M}$, which represents the predicted residual for individual $i$ if they belonged to location $m$. Recall that we define the actual residual as $R_i^g = Y_i - \hat{f}_{\text{base}}(X_i,g)$.
To find a suitable cluster for location $g$, we solve the following optimization problem:
\begin{equation} \label{eq:weights_generation}
\begin{aligned}
\min_{\mathbf{z} \in \{0,1\}^{|\mathcal{M}|}} \quad 
    & \sum_{i=1}^{|\mathcal{D}^{g,20}_{\text{train}}|}
      \left(
        R^g_i -
        \frac{\sum_{m=1}^{|\mathcal{M}|} z_m r_{im} n_m}
             {\sum_{m=1}^{|\mathcal{M}|} z_m n_m}
      \right)^2 \\
\text{s.t.} \quad 
    & z_g = 1,\quad \sum_{m=1}^{|\mathcal{M}|} z_m \le \lambda.
\end{aligned}
\end{equation}
where $z_m$ is a binary decision variable indicating whether location $m$ is included in the cluster for location $g$, and $n_m$ is the size of location $m$. We denote the selected cluster as $\mathcal{C}(g) = \{ m : z^*_m = 1 \}$, where $\mathbf{z}_g^*$ is the optimal solution to Problem \ref{eq:weights_generation}. The objective seeks to find a weighted combination of residual models that best approximates the actual residuals of location $g$, giving more weight to locations with larger sample sizes. While many current methods for data selection focus on data weighting rather than pooling \cite{idrissi2022simpledatabalancingachieves,xie_doremi_2023,jeong2025outofdistributiongeneralizationrandomdense}, we found that many state-of-the-art algorithms perform sub-optimally under re-weighting for real-world data, which we also show with one of our benchmarks (RWG). Thus, we focus on data pooling instead of data weighting.

Problem~\ref{eq:weights_generation} is nonlinear and nonconvex, but it can be optimally solved using modern mixed-integer programming solvers. The pipeline is summarized in Algorithm \ref{alg:xlearn_clustering}. Inspired by stabilization procedures in \citet{meinshausen2010stability}, we repeat this procedure $\gamma$ times with different training splits (see Appendix \ref{clustered_x_param} for all hyperparameter information). As a post-processing step, we then define $\mathbf{w}_g \in [0,1]^{|\mathcal{M}|}$ which denotes a vector where each entry $w_{g,m}$ represents the fraction of iterations in which $z^*_{g,m} = 1$ when solving Problem \ref{eq:weights_generation} for location $g$. Since solving Problem~\ref{eq:weights_generation} hundreds of times for each location can be computationally expensive when $M$ is large, we restrict the candidate set to a random subset of locations always including $g$ in each run (denoted by a sparsity-regularization parameter $\lambda$).

\begin{algorithm}[tb]
\caption{CTRL: Generate Decision Variable}
\label{alg:xlearn_clustering}
\begin{algorithmic}[1]

\STATE {\bfseries Input:} $\mathcal{D}_{\text{train}}, g, \text{regularization param }\lambda, \text{seed}\leq\gamma$
\STATE {\bfseries Output:} Weight vector $\mathbf{z}^*_g$ for location $g$

\STATE Split $\mathcal{D}_{\text{train}}$ into $\mathcal{D}_{\text{train}}^{80}$ (train) and $\mathcal{D}_{\text{train}}^{20}$ (validation)

\STATE Train $\hat{f}_{\text{base}}$ on $\mathcal{D}_{\text{train}}^{80}$

\FOR{each group $m \in \mathcal{M}$}
    \STATE Compute $R^m_i = Y_i - \hat{f}_{\text{base}}(X_i,m)$ for $i \in \mathcal{D}_{\text{train}}^{m,80}$
    \STATE Train residual model $\hat{f}^m_{\text{residual}}$ on $\mathcal{D}_{\text{train}}^{m,80}$
    \STATE Compute $r_{im} = \hat{f}^m_{\text{residual}}(X_i)$ for $i \in \mathcal{D}_{\text{train}}^{g,20}$
\ENDFOR

\STATE Compute $R^g_i = Y_i - \hat{f}_{\text{base}}(X_i,g)$ for $i \in \mathcal{D}_{\text{train}}^{g,20}$

\STATE Solve Problem~\ref{eq:weights_generation} using $R^g$ and $r_m$ to obtain $\mathbf{z}^*$

\STATE {\bfseries return:} $\mathbf{z}^*_g$

\end{algorithmic}
\end{algorithm}

\begin{algorithm}[ht!]
\caption{CTRL: Get Optimal Cluster}
\label{alg:cluster_selection}
\begin{algorithmic}[1]

\STATE {\bfseries Input:} $\mathcal{D}_{\text{train}}, g, \mathbf{w}_{g}, \gamma$
\STATE {\bfseries Output:} Optimal cluster for location $g$

\FOR{$\text{iter} = 1$ {\bfseries to} $\gamma$}

    \STATE Split $\mathcal{D}_{\text{train}}$ into 
        $\mathcal{D}_{\text{train}}^{80}$ (train) and 
        $\mathcal{D}_{\text{train}}^{20}$ (validation)

    \STATE Train base model $\hat{f}_{\text{base}}$ on $\mathcal{D}_{\text{train}}^{80}$

    \FOR{$k = 1$ {\bfseries to} $K$}

        \STATE $C_k(g) \gets$ top-$k$ indices of $\mathbf{w}_{g}$

        \STATE Train $\hat{f}_{\text{residual}}^{C_k(g)}$ 
        on $\mathcal{D}_{\text{train}}^{C_k(g),80}$

        \STATE Compute predicted residuals $r_{ig}$ 
        on $\mathcal{D}_{\text{train}}^{g,20}$

        \STATE $\displaystyle
        \text{MSE}_{g}^{(k)} = 
        \frac{1}{n_g}
            \sum_{i \in \mathcal{D}_{\text{train}}^{g,20}}
            \left(
                Y_i^g -
                \left(
                    \hat{f}_{\text{base}}(X_i,g) + r_{ig}
                \right)
            \right)^2$
    \ENDFOR

\ENDFOR

\STATE Compute $\bar{\text{MSE}}_{g}^{k}$ and $\text{SE}_{\text{MSE},g}^{k}$ for $k = 1,\ldots,10$

\STATE $k_{\min} = \arg\min_{k} \bar{\text{MSE}}_{g}^{k}$

\STATE $\text{MSE}_{g}^{\text{cutoff}}
        = \bar{\text{MSE}}_{g}^{k_{\min}}
        + \text{SE}_{\text{MSE},g}^{k_{\min}}$

\STATE $k_{g}^{*} = 
        \min\left\{
            k : \bar{\text{MSE}}_{g}^{k} 
            \le \text{MSE}_{g}^{\text{cutoff}}
        \right\}$

\STATE {\bfseries return} top-$k_{g}^{*}$ locations from $\mathbf{w}_{g}$

\end{algorithmic}
\end{algorithm}

We use Algorithm~\ref{alg:cluster_selection} to construct the final cluster for location $g$. We begin with $\mathcal{C}(g) = \{g\}$ and iteratively add locations in decreasing order of values in $\mathbf{w}_g$. We consider the top $K$ locations in $\mathbf{w}_g$. Experimentally we see that $K=10$ is sufficient (Figure \ref{fig:top_k}). After each addition, we train $\hat{f}_{\text{residual}}^{\mathcal{C}(g)}$ and evaluate its MSE on $\mathcal{D}_{\text{train}}^{g,20}$. We repeat across $\gamma$ splits and apply the ``1 Standard Error Rule'' \cite{1se} to determine the optimal cluster for $g$, $\mathbf{C^*(g)}$.  Lastly, we train $\hat{f}_{\text{base}}$ on $\mathcal{D}_{\text{train}}$, train $\hat{f}_{\text{residual}}^{\mathbf{C^*(g)}}$ on $\mathcal{D}_{\text{train}}^{\mathbf{C^*(g)}}$, and add them to model $\hat{f}_{\text{CTRL}}$ which we use to evaluate on $\mathcal{D}^g_{\text{test}}$. Repeat Alg \ref{alg:cluster_selection} for all $g \in \mathcal{M}$. We include Appendix \ref{qual_analysis} to build more intuition on the generated clusters.

\section{Theoretical Results} \label{sec:theory}

We now provide theoretical support for CTRL's objective. Proposition~\ref{prop:leafwise-ctrl-risk} shows that, for leafwise-constant function classes, minimizing the CTRL prediction risk is asymptotically equivalent to optimizing over convex combinations of source-specific residual fits. The result relies on the assumption that there is negligible shift in covariates across sources at the leaf level. In the Swiss asylum seeker setting, this assumption is well motivated: the historical assignment of asylum seekers to cantons was approximately random. 

\begin{proposition}[Link between residual optimization and CTRL risk]\label{prop:leafwise-ctrl-risk}
Let $\mathcal{F}$ be the class of functions that are constant on leaves $L \in \mathcal{L}$, where $\mathcal{L}$ is a finite partition of $\mathcal{X}$ (e.g., as in regression trees with fixed splits). Let $\mathcal{F}'$ be a class of functions that are constant on leaves $B\in\mathcal{B}$, where $\mathcal{B}$ is a finite partition of $\mathcal{X}\times\mathcal{M}$. Assume there is negligible covariate shift across sources at the leaf level, i.e., for each $L\in\mathcal L$, $\sup_m\big|\hat P_m(L)- P_1(L)\big|=o_p(1)$, and that outcomes are bounded ($|Y|\le C_Y$). In the following, $E_m[\cdot]$ denotes population expectation under source $m$, $\hat E_m[\cdot]$ denotes empirical mean over source-$m$ samples, and $\hat E_{\text{pooled}}[\cdot]$ denotes empirical mean over pooled samples from all locations. Define
\begin{equation*}
\hat f_{\text{base}}(\cdot)\in\arg\min_{f \in \mathcal{F}'} \hat E_{\text{pooled}}[(Y-f(X,M))^2], \, R^m:=Y-\hat f_{\text{base}}(X,m).
\end{equation*}
Fix weights $w\in\Delta^{|\mathcal M|}:=\{w_m\ge0,\sum_m w_m=1\}$  and define
\begin{equation*}
\hat f_{\text{residual}}^w(\cdot)\in\arg\min_{f_\text{residual} \in \mathcal{F}} \sum_m w_m\hat E_m[(R^m-f_{\text{residual}}(X))^2],
\end{equation*}
with final predictor
\begin{equation*}
\hat f_{\mathrm{CTRL}}^w(x,g):=\hat f_{\text{base}}(x,g)+\hat f_{\text{residual}}^w(x).
\end{equation*}
Here, the weights $w$ are later optimized to improve performance on the target $g$. Finally, for each source $m\in\mathcal M$, define
\begin{equation*}
r^m(\cdot) \in\arg\min_{f_\text{residual} \in \mathcal{F}} \hat E_m[(R^m-f_\text{residual}(X))^2].
\end{equation*}
Then, for any fixed target source $g$,
\begin{equation*}
E_g\!\left[(Y-\hat f_{\mathrm{CTRL}}^w(X,g))^2\right]
=
E_g\!\left[\left(R^g-\sum_m w_m r^m(X)\right)^2\right]+o_p(1).
\end{equation*}
\end{proposition}
The proof is deferred to Appendix~\ref{app:proof-leafwise}. In Problem~\eqref{eq:weights_generation}, CTRL selects a cluster via binary inclusion variables $z_m\in\{0,1\}$. For any fixed $z$ with at least one active source, define the induced simplex weights
\begin{equation*}
w_m(z):=\frac{z_m n_m}{\sum_j z_j n_j},
\qquad \sum_m w_m(z)=1.
\end{equation*}
Thus, each discrete cluster corresponds to a point in the simplex used in Proposition~\ref{prop:leafwise-ctrl-risk}. The proposition analyzes the continuous weighted objective; the algorithmic search over $z$ can be viewed as searching over a structured subset of those weights (with optional stability-aggregation across splits in the empirical procedure).

Proposition~\ref{prop:leafwise-ctrl-risk} justifies CTRL's strategy of searching over cluster compositions: since the prediction risk decomposes into a combination of source-wise residual fits, optimizing over which sources to include directly targets the quantity of interest. 

In Appendix~\ref{sec:excess-risk}, we complement this result with an excess risk bound under a random distributional shift model (Proposition~\ref{prop:excess-risk}), which characterizes the tradeoff between  variance reduction from pooling additional data and error introduced by distribution shift.

\section{Runtime}\label{sec:runtime}
Let $N_g = |\mathcal{D}^{g,20}_{\text{train}}|$ denote the number of 
observations used in Equation \ref{eq:weights_generation} for location $g$. In each iteration, CTRL solves the mixed-integer
problem in Problem~\eqref{eq:weights_generation} over all $|\mathcal{M}|$
locations. For any given $\mathbf{z}$, at most $\lambda$ locations contribute
to the prediction, so evaluating the objective costs $O(N_g \lambda)$. The
feasible set consists of all subsets of locations of size at most $\lambda$,
i.e., $\sum_{k \le \lambda} \binom{|\mathcal{M}|-1}{k-1}$ possibilities, which
is combinatorial but still far smaller than the full $2^{|\mathcal{M}|}$ space
when $\lambda$ is small. Modern mixed-integer solvers exploit this structure
via branch-and-bound, so the effective search is typically far below the worst
case in the many-sources regimes we consider (dozens of locations and small
$\lambda$). Stability selection repeatedly solves this constrained problem on resampled
data and aggregates the results, providing a practical way to identify
locations that are stably useful for transfer without requiring an exhaustive
search. The $\gamma$ runs are independent and therefore
\emph{fully parallelizable}, yielding near-linear speedups with additional
cores. In our experiments, CTRL is computationally feasible (e.g., roughly
three hours on an 8-core machine for $\sim 25{,}000$ observations), and
$\gamma$ can be relatively small—Figure~\ref{fig:top_5_metric} shows
performance stabilizing after approximately 100 iterations. Additional speedups
are possible through continuous relaxations, greedy residual-similarity
pruning, or using smaller values of $\lambda$ and appropriately tuning $\gamma$
for larger $|\mathcal{M}|$. Lastly, once trained, CTRL incurs no additional inference-time cost relative to TRL.

\section{Datasets}
We evaluate our method on 5 datasets and provide 4 in our codebase. See Table \ref{tab:dataset-summary} for dataset sizes, features, source and outcome descriptions, etc.

\paragraph{Synthetic dataset.} We generate a dataset of 40,000 individuals divided into 50 locations, ranging from 40 to 2,000 in size. For each row, we compute an outcome probability using a weighted combination of a global linear model (30\%) and location-specific local linear models (70\%). A subset of locations share similar local models to simulate a clustered structure. Binary outcomes are sampled from these probabilities, enabling evaluation on both the prediction and ranking tasks. Implementation details in Appendix: \ref{synthetic_details}.
\paragraph{Swiss asylum seekers dataset.} 
This administrative dataset from the Central Migration Information System (ZEMIS) of the Swiss State Secretariat for Migration contains background characteristics, assigned locations (i.e. cantons, of which there are 26), and employment outcomes for adult asylum seekers entering Switzerland who were granted a protected status.\footnote{Specifically, the data contain asylum seekers who eventually received full protection status specified under the Geneva Convention as well as those whose claim for Geneva protection status was rejected but were awarded subsidiary protection.} The dataset comprises individuals who were assigned to a canton between 2018-2022. We use a binary two-year employment outcome as the outcome of interest, encoding whether an individual found \emph{any} employment in their first two years after assignment. This dataset is not publicly available but can be requested from the Swiss State Secretariat for Migration. The study received IRB approval from Harvard University (IRB22-1083).

\paragraph{Education dataset.} We use U.S. Census data from 2014 to predict whether individuals over 18 have completed high school -- using \cite{ding2021retiring} for preprocessing. We select 14 demographic and background variables (e.g., language, disability status, state, country of birth), and use state as the location variable. 
\paragraph{Dissecting Bias health dataset.} This semi-synthetic health dataset \cite{obermeyer} includes patient records with binary outcomes marking chronic illness. Each observation includes features like demographics, comorbidities, prior costs, biomarkers, and medications. Instead of geographic locations as the source, we define ``source'' as combinations of race, age, and gender  \textit{(e.g. Black, 20-30 years old, female)}.

\paragraph{UK asylum decisions dataset.} This dataset\footnote{https://www.gov.uk/government/statistical-data-sets/asylum-and-resettlement-datasets} includes asylum applications and resettlement decisions in the United Kingdom, disaggregated by nationality, demographic attributes, and administrative attributes such as time and case details. We define ``sources'' as nationalities (instead of locations) and predict the likelihood of an individual asylum request being approved by a judge.

\section{Model Evaluation}
\subsection{Metrics}
We evaluate models using three key metrics:  
(1) the average outcome among the top 20\%\footnote{We chose the 20\% cutoff for these results to balance identifying top performers while ensuring location-level variation in candidate selection. We provide results for other cutoffs in the appendix and our model still outperforms the others (Appendix Table \ref{tab:combined-rwa-all-clean}).} of individuals ranked by each location according to each method, or Rank-Weighted Average (RWA),
(2) overall mean squared error (MSE), and
(3) MSE for small locations (defined as the bottom third by location size).
The RWA, inspired by \citet{Yadlowsky02012025}, evaluates whether a model effectively prioritizes individuals who are likely to benefit most from being matched to a given location. For each location \( g \in \mathcal{M} \), we identify the top 20\% of individuals with the highest \( \hat{Y}_{ig} \), where \( \hat{Y}_{ig} \) denotes the prediction for individual \( i \) at location \( g \). Let \( S_g \) be the set of these top-scoring individuals for location \( g \).

\[
\text{RWA} = \frac{1}{|\mathcal{A}|} \sum_{i \in \mathcal{A}} Y_i \:\:\:\:\: \text{where} \:\:\:\:\: \mathcal{A} = \bigcup_{g \in \mathcal{M}'} \left\{ i \in S_g : M_i = g \right\}
\]

The set \( \mathcal{A} \) contains individuals who were both in the top 20\% for some location \( g \)  and actually assigned to \( g \) in the data. Rather than consider all locations $\mathcal{M}$, we restrict attention to $\mathcal{M}'$---the set of locations where $|\left\{ i \in S_g : M_i = g \right\}|\geq 10$ across all models.

In the Swiss asylum seeker context, the historical assignment procedure satisfies conditional ignorability\footnote{conditional upon the covariates we have access to}, so the predictions can be interpreted causally. Thus, in this setting RWA approximates the counterfactual outcome if the top 20\% of individuals were actually assigned to $g$. RWA is the most relevant evaluation metric in this context, as it aligns with our use case of improving downstream assignment outcomes rather than predictive accuracy in isolation. Lastly, all results are averaged over 10 splits, highlighting the robustness of CTRL’s improvements.

\subsection{Benchmarks}
We evaluate against several benchmark algorithms and underlying architectures. Following prior work \cite{NEURIPS2023_a134eaeb, bansak2024learning}, we identify tree ensembles as the most consistently high-performing model architectures for our datasets. As such, we include results for Bayesian Additive Regression Trees (BART) and Random Forest (RF), which serve as strong performance benchmarks for our setting. Given the real-world settings in our work, we also report results using  interpretable models (linear regression and decision trees) which are often preferred by practitioners in high-stakes contexts \cite{rudin2019stopexplainingblackbox}.

For benchmarks, we compare CTRL against global and local models, which correspond to full pooling and no pooling baselines, respectively. We also include transfer residual learning (TRL) to isolate the effect of adaptive clustering. We also include two state-of-the-art methods for learning with imbalanced or heterogeneous-source data: JTT and RWG \cite{liu2021justtraintwiceimproving,idrissi2022simpledatabalancingachieves}. For the  RWG benchmark, we assign weights to ensure that each location contributes equally to the model’s objective, so that the loss from a small location is given the same importance as that from a large location. While Group-DRO is another benchmark often used for predictions for datasets with multiple subgroups, it is irrelevant for our use case because it would output the same predictions for each location, which does not address our goal of differentiating between locations. 

Finally, we conduct an in-depth comparison of clusters discovered by CTRL against two simpler baselines: one that clusters locations using the Wasserstein distance between residual distributions, and another based on correlations between predicted residuals. Full implementation details are provided in Appendix~\ref{simpleclustering}.

\section{Experimental Results}

\subsection{Rank Weighted Average (RWA)}
We begin by evaluating the RWA produced by CTRL relative to our benchmarks (Table~\ref{tab:rwa_only}). For this analysis, we use the Synthetic, Swiss Asylum, and Education\footnote{We include an Educational Outcomes dataset to demonstrate the robustness of CTRL. Although this dataset does not permit causal interpretation—because individuals are not randomly assigned to states—it provides a useful testbed for assessing generalization behavior.} datasets, as each contains an assignable location variable. RWA is the most relevant metric for our real-world deployment in the Swiss Asylum Seeker application, where the goal is to identify individuals who are the best match for each geographic location. This capability is essential because asylum system staff---supported by the algorithm---make location assignments that have meaningful downstream consequences.

Across all three datasets, CTRL consistently achieves the highest average RWA, and in the few instances where it is not the top performer, its performance is comparable to the best alternative. Crucially, CTRL’s advantage over the global model indicates that it is not merely ranking the most “universally employable’’ individuals highest across all locations. Instead, CTRL captures location-specific heterogeneity, adapting rankings to the distinctive labor markets, constraints, and opportunities associated with each location. This demonstrates that CTRL learns meaningful structure that improves decision quality in settings where local variation matters.

\begin{table}[ht!]
\centering
\caption{Average RWA across 3 datasets (top 20\%). Higher is better. Standard errors in parentheses.}
\small
\resizebox{\columnwidth}{!}{
\setlength{\tabcolsep}{2pt}  
\begin{tabular}{p{1.1cm} p{0.9cm} | cccc}
\toprule
\textbf{Dataset} & \textbf{Model} & Reg & Tree & RF & BART \\
\midrule
\multirow{6}{=}{\raggedright Synthetic} & JTT & 0.783 (0.001) & 0.618 (0.005) & 0.777 (0.001) & 0.798 (0.001) \\
 & RWG & 0.783 (0.001) & 0.683 (0.005) & 0.765 (0.001) & 0.775 (0.001) \\
 & Global & 0.783 (0.001) & 0.705 (0.003) & 0.770 (0.001) & 0.819 (0.001) \\
 & Local & 0.958 (0.001) & 0.690 (0.004) & 0.912 (0.001) & 0.926 (0.002) \\
 & TRL & 0.958 (0.001) & 0.746 (0.002) & 0.896 (0.001) & 0.929 (0.001) \\
 & CTRL & \textbf{0.962 (0.004)} & \textbf{0.772 (0.010)} & \textbf{0.913 (0.001)} & \textbf{0.939 (0.001)} \\
\midrule
\multirow{6}{=}{\raggedright Swiss Asylum Seekers} & JTT & 0.288 (0.004) & 0.230 (0.004) & 0.391 (0.004) & 0.363 (0.004) \\
 & RWG & 0.296 (0.004) & 0.322 (0.005) & 0.384 (0.005) & 0.362 (0.004) \\
 & Global & 0.295 (0.005) & 0.331 (0.004) & 0.394 (0.004) & 0.367 (0.004) \\
 & Local & 0.297 (0.005) & 0.306 (0.004) & 0.375 (0.004) & 0.346 (0.005) \\
 & TRL & 0.310 (0.004) & 0.345 (0.004) & 0.400 (0.004) & \textbf{0.373 (0.004)} \\
 & CTRL & \textbf{0.312 (0.004)} & \textbf{0.348 (0.003)} & \textbf{0.401 (0.004)} & 0.372 (0.004) \\
\midrule
\multirow{6}{=}{\raggedright Education} & JTT & 0.937 (0.000) & 0.916 (0.005) & 0.907 (0.001) & \textbf{0.946 (0.000)} \\
 & RWG & 0.939 (0.000) & 0.885 (0.016) & 0.941 (0.000) & 0.941 (0.000) \\
 & Global & 0.940 (0.000) & 0.854 (0.017) & 0.943 (0.000) & 0.944 (0.000) \\
 & Local & 0.898 (0.014) & 0.875 (0.002) & 0.931 (0.000) & 0.942 (0.000) \\
 & TRL & \textbf{0.941 (0.000)} & 0.927 (0.001) & 0.945 (0.000) & 0.944 (0.000) \\
 & CTRL & \textbf{0.941 (0.000)} & \textbf{0.931 (0.001)} & \textbf{0.946 (0.000)} & 0.944 (0.000) \\
\bottomrule
\end{tabular}}
\label{tab:rwa_only}
\end{table}

\subsection{MSE Results}
We next examine predictive accuracy using MSE across all datasets. In addition to the datasets discussed earlier for RWA, this analysis includes the UK Asylum Decisions and Dissecting Bias in Health datasets, where “sources’’ are defined by demographic attributes rather than geographic locations. Because these datasets do not support assignable source attributes, RWA is not applicable; however, they provide valuable testbeds for assessing CTRL’s behavior in imbalanced, many-source prediction settings.

\begin{table*}[ht!]
\centering
\caption{Average MSE and Small MSE averaged over 10 splits for all datasets. Standard Error in parentheses.}
\footnotesize
\setlength{\tabcolsep}{2.5pt}
\resizebox{\textwidth}{!}{
\begin{tabular}{p{1.3cm} p{1.1cm} | cccc | cccc}
\toprule
\textbf{Dataset} & \textbf{Model} &
\multicolumn{4}{c}{\textbf{MSE}} &
\multicolumn{4}{c}{\textbf{Small MSE}} \\
& & Reg & Tree & RF & BART & Reg & Tree & RF & BART \\
\midrule
\multirow{6}{=}{\raggedright Synthetic} & JTT & 0.209 (0.000) & 0.256 (0.002) & 0.222 (0.000) & 0.246 (0.001) & 0.227 (0.003) & 0.263 (0.003) & 0.232 (0.001) & 0.261 (0.003) \\
 & RWG & 0.209 (0.000) & 0.244 (0.002) & 0.219 (0.000) & 0.214 (0.000) & 0.227 (0.003) & 0.267 (0.003) & 0.228 (0.002) & 0.227 (0.003) \\
 & Global & 0.209 (0.000) & 0.233 (0.001) & 0.215 (0.000) & 0.198 (0.000) & 0.227 (0.002) & 0.249 (0.002) & 0.228 (0.001) & 0.230 (0.001) \\
 & Local & \textbf{0.132 (0.000)} & 0.245 (0.001) & 0.191 (0.000) & 0.157 (0.000) & 0.164 (0.002) & 0.269 (0.002) & 0.212 (0.000) & 0.188 (0.001) \\
 & TRL & 0.137 (0.000) & 0.228 (0.001) & 0.176 (0.000) & 0.154 (0.000) & 0.160 (0.001) & 0.250 (0.002) & 0.199 (0.001) & 0.182 (0.001) \\
 & CTRL & 0.136 (0.002) & \textbf{0.220 (0.003)} & \textbf{0.170 (0.001)} & \textbf{0.150 (0.001)} & \textbf{0.146 (0.004)} & \textbf{0.236 (0.003)} & \textbf{0.182 (0.001)} & \textbf{0.167 (0.002)} \\
\midrule
\multirow{6}{=}{\raggedright Swiss Asylum Seekers} & JTT & 0.108 (0.001) & 0.204 (0.001) & 0.138 (0.000) & 0.104 (0.001) & 0.107 (0.004) & 0.222 (0.006) & 0.144 (0.001) & 0.101 (0.004) \\
 & RWG & 0.107 (0.001) & 0.107 (0.001) & 0.097 (0.001) & 0.100 (0.001) & \textbf{0.105 (0.004)} & 0.108 (0.003) & 0.095 (0.003) & 0.097 (0.003) \\
 & Global & 0.108 (0.001) & 0.104 (0.001) & 0.096 (0.001) & 0.099 (0.001) & 0.106 (0.004) & \textbf{0.102 (0.003)} & 0.095 (0.003) & \textbf{0.096 (0.003)} \\
 & Local & 0.108 (0.001) & 0.106 (0.001) & 0.097 (0.001) & 0.103 (0.001) & 0.115 (0.006) & 0.116 (0.004) & 0.102 (0.003) & 0.108 (0.003) \\
 & TRL & \textbf{0.106 (0.001)} & 0.102 (0.001) & \textbf{0.092 (0.001)} & \textbf{0.097 (0.001)} & 0.106 (0.003) & 0.103 (0.003) & \textbf{0.094 (0.003)} & \textbf{0.096 (0.003)} \\
 & CTRL & \textbf{0.106 (0.001)} & \textbf{0.101 (0.001)} & \textbf{0.092 (0.001)} & 0.098 (0.001) & 0.107 (0.003) & \textbf{0.102 (0.003)} & \textbf{0.094 (0.003)} & \textbf{0.096 (0.003)} \\
\midrule
\multirow{6}{=}{\raggedright Education} & JTT & 0.153 (0.000) & 0.227 (0.002) & 0.218 (0.000) & 0.197 (0.000) & 0.156 (0.001) & 0.232 (0.004) & 0.213 (0.001) & 0.202 (0.001) \\
 & RWG & 0.153 (0.000) & 0.161 (0.001) & 0.154 (0.000) & 0.152 (0.000) & 0.155 (0.001) & 0.162 (0.001) & 0.153 (0.000) & \textbf{0.152 (0.000)} \\
 & Global & 0.152 (0.000) & 0.161 (0.001) & 0.151 (0.000) & 0.151 (0.000) & \textbf{0.154 (0.000)} & 0.163 (0.001) & 0.153 (0.000) & 0.153 (0.000) \\
 & Local & 0.171 (0.004) & 0.173 (0.000) & 0.166 (0.000) & 0.151 (0.000) & 0.171 (0.002) & 0.179 (0.001) & 0.167 (0.000) & 0.159 (0.001) \\
 & TRL & \textbf{0.151 (0.000)} & 0.156 (0.001) & \textbf{0.150 (0.000)} & \textbf{0.150 (0.000)} & 0.155 (0.000) & 0.163 (0.001) & 0.153 (0.000) & 0.154 (0.000) \\
 & CTRL & \textbf{0.151 (0.000)} & \textbf{0.154 (0.000)} & \textbf{0.150 (0.000)} & \textbf{0.150 (0.000)} & \textbf{0.154 (0.000)} & \textbf{0.160 (0.001)} & \textbf{0.152 (0.000)} & 0.153 (0.000) \\
\midrule
\multirow{6}{=}{\raggedright Dissecting Health Bias} & JTT & 0.167 (0.001) & 0.228 (0.001) & 0.200 (0.000) & 0.195 (0.000) & 0.140 (0.001) & 0.202 (0.004) & 0.173 (0.001) & 0.173 (0.002) \\
 & RWG & \textbf{0.166 (0.000)} & 0.175 (0.001) & 0.250 (0.001) & 0.165 (0.001) & 0.138 (0.001) & 0.150 (0.002) & 0.200 (0.004) & \textbf{0.136 (0.002)} \\
 & Global & \textbf{0.166 (0.000)} & 0.173 (0.001) & 0.166 (0.001) & 0.166 (0.001) & \textbf{0.137 (0.001)} & 0.153 (0.002) & 0.139 (0.002) & \textbf{0.136 (0.002)} \\
 & Local & 0.174 (0.002) & 0.192 (0.000) & 0.172 (0.000) & 0.166 (0.000) & 0.151 (0.002) & 0.165 (0.003) & 0.144 (0.002) & 0.144 (0.002) \\
 & TRL & \textbf{0.166 (0.000)} & 0.171 (0.001) & \textbf{0.165 (0.001)} & \textbf{0.164 (0.000)} & \textbf{0.137 (0.001)} & \textbf{0.149 (0.002)} & \textbf{0.135 (0.001)} & 0.137 (0.001) \\
 & CTRL & \textbf{0.166 (0.000)} & \textbf{0.170 (0.001)} & \textbf{0.165 (0.001)} & \textbf{0.164 (0.000)} & \textbf{0.137 (0.001)} & \textbf{0.149 (0.002)} & 0.136 (0.001) & \textbf{0.136 (0.001)} \\
\midrule
\multirow{6}{=}{\raggedright UK Asylum Decisions} & JTT & 0.190 (0.000) & 0.235 (0.001) & 0.222 (0.000) & 0.202 (0.000) & 0.147 (0.001) & 0.209 (0.002) & 0.190 (0.001) & 0.153 (0.002) \\
 & RWG & 0.188 (0.000) & 0.194 (0.000) & 0.193 (0.000) & 0.186 (0.000) & \textbf{0.142 (0.001)} & \textbf{0.148 (0.001)} & 0.148 (0.001) & \textbf{0.140 (0.001)} \\
 & Global & 0.188 (0.000) & 0.195 (0.000) & 0.193 (0.000) & 0.185 (0.000) & 0.143 (0.001) & 0.154 (0.001) & 0.153 (0.001) & 0.144 (0.001) \\
 & Local & 0.197 (0.001) & 0.197 (0.000) & 0.191 (0.000) & 0.187 (0.000) & 0.151 (0.002) & 0.156 (0.002) & 0.143 (0.001) & 0.145 (0.001) \\
 & TRL & \textbf{0.187 (0.000)} & 0.191 (0.000) & 0.193 (0.000) & \textbf{0.183 (0.000)} & 0.143 (0.002) & \textbf{0.148 (0.002)} & \textbf{0.141 (0.001)} & 0.141 (0.001) \\
 & CTRL & \textbf{0.187 (0.000)} & \textbf{0.188 (0.000)} & \textbf{0.190 (0.000)} & \textbf{0.183 (0.000)} & 0.143 (0.001) & 0.149 (0.001) & \textbf{0.141 (0.001)} & 0.141 (0.001) \\
\bottomrule
\end{tabular}}
\label{tab:all_mse_smallmse}
\end{table*}

Although aggregate predictive accuracy is not the primary objective of this work, MSE remains an important diagnostic. We report MSE for transparency and to confirm that CTRL’s decision advantages do not come at the cost of degraded prediction performance. Ideally, CTRL should maintain or improve predictive accuracy while enabling superior downstream decisions. As shown in Table~\ref{tab:all_mse_smallmse}, we see this effect: CTRL consistently matches or outperforms all benchmark models on overall MSE. In cases where it is not the top performer, it remains very close to the best alternative.

On the Synthetic data, the data-generating process is linear at the location level, so the local regression model performs strongly on MSE. However, CTRL achieves comparable accuracy to the local approach, unlike the alternatives. Moreover, the gap between CTRL and the baselines becomes more pronounced on Small MSE, illustrating a key weakness of local models: they perform poorly for small-source groups due to limited data. CTRL avoids this pitfall and delivers stable accuracy across both large and small locations.

For a few datasets, RWG likely performs well on small MSE because it explicitly upweights data from small locations. However, this improvement with substantially worse overall MSE and worse RWA, making RWG less suitable for practical deployment. CTRL, in contrast, achieves small MSE that is competitive with the strongest baselines while  delivering substantially higher RWA. This demonstrates that CTRL remains robust even in data-sparse regions without compromising downstream utility.

\subsection{Overall Performance}
\begin{figure}[ht!]
  \centering
  \includegraphics[width=0.8\linewidth]{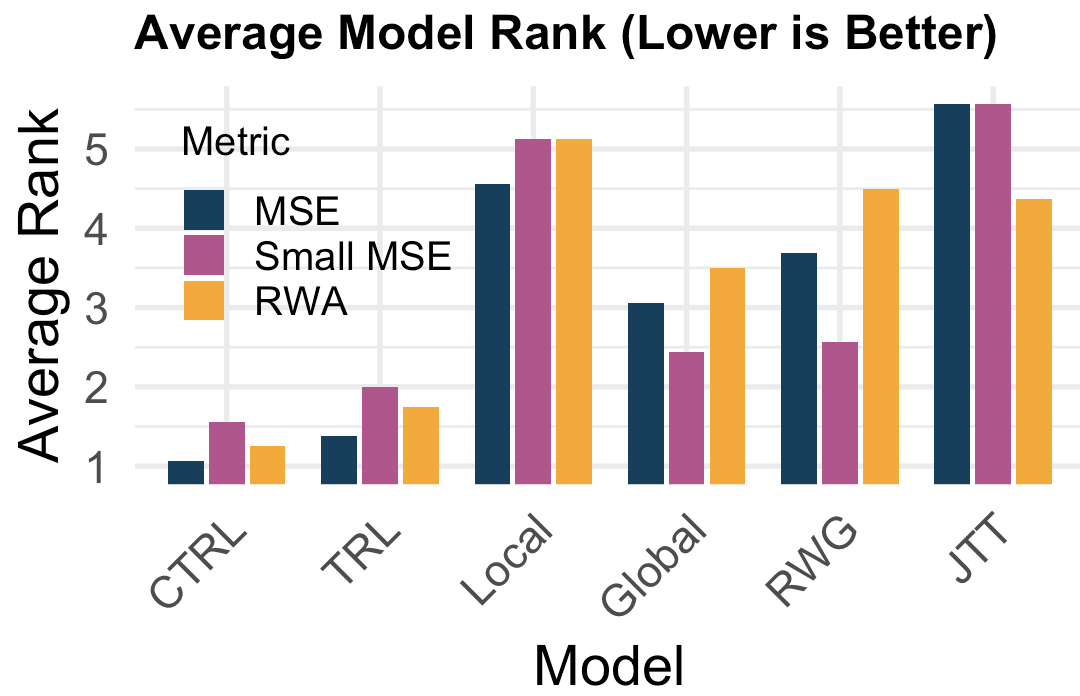}
  \caption{Average model performance ranks across datasets.}
  \label{fig:avg_performance}
  \vspace{-0.4cm}
\end{figure}

To compare overall model performance across datasets, we use an average rank metric which is standard for multi-dataset evaluation \cite{10.5555/1248547.1248548}. We exclude the Synthetic dataset since it was constructed specifically for our use case. Figure~\ref{fig:avg_performance} reports the average rank of each model across all datasets for all available evaluation metrics for those datasets. CTRL shows the best rank against benchmarks. Additional aggregated results are in the appendix.

\subsection{Clustering Analysis} \label{sec:clusteranalysis}
To compare alternative distance strategies for measuring similarity between groups, we leverage the synthetic dataset, for which the true cluster assignments are known by construction. This setting allows us to directly evaluate how well a given distance metric recovers the underlying cluster structure, which is not observable in real-world datasets. Our analysis compares three approaches: CTRL’s learned distance measure (obtained by applying Eqn.~\ref{eq:weights_generation} across $\gamma$ splits), a Wasserstein distance baseline, and a correlation-based distance computed between group-level residual vectors. The Wasserstein distance captures distributional similarity between residuals, whereas the correlation-based distance captures similarity in residual patterns. We define the following evaluation metric.
\paragraph{Weighted Precision@3.}
Let $g \in \{1,\dots,|\mathcal{M}|\}$ index groups, and let $C(g)$ denote the true cluster label of group $g$. Let $m_r(g)$ denote the group ranked $r$-th closest to $g$ under the distance metric. We define a weighted precision metric over the top three nearest neighbors as
\begin{equation}
\label{eq:weighted_precision}
\text{WP@3}
\;=\;
\frac{
\sum_{g=1}^{|\mathcal{M}|}
\sum_{r=1}^{3}
\frac{1}{r}
\cdot
\mathbb{I}\!\left\{ C\!\left(m_r(g)\right) = C(g) \right\}
}{
|\mathcal{M}| \cdot \sum_{r=1}^{3} \frac{1}{r}
}.
\end{equation}

This metric assigns greater weight to closer neighbors, reflecting the intuition that correct matches at smaller ranks are more informative. We focus on the top three nearest groups, which corresponds to the average cluster size in the synthetic data. The score is normalized by the maximum achievable WP@3 value to ensure comparability across settings. 

Table~\ref{tab:distance_comparison} reports results averaged over 10 splits. The proposed CTRL distance substantially outperforms the Wasserstein and correlation baselines, achieving an average weighted precision of 83\%, indicating more reliable recovery of true cluster neighbors. These results illustrate that generic distributional or shape-based distances can fail to reflect predictive relevance under heterogeneous distribution shift, demonstrating a need for methods such as CTRL.

\begin{table}[ht!]
\centering
\caption{Weighted Precision@3 on synthetic dataset. Higher values indicate better recovery of true cluster structure.}
\small
\begin{tabular}{l c}
\toprule
\textbf{Distance Metric} & \textbf{Weighted Precision@3} \\
\midrule
Correlation & 0.067 (0.003) \\
Wasserstein Distance & 0.309 (0.029) \\
CTRL  & \textbf{0.832 (0.009)} \\
\bottomrule
\end{tabular}
\label{tab:distance_comparison}
\end{table}

\section{Discussion and Conclusion}
In the presence of distribution shift across many datasets of varying sizes, standard transfer learning approaches can struggle due to uneven sample sizes and variable outcome distributions. We propose CTRL, which simultaneously leverages the richness of the full dataset while learning location-specific heterogeneity. CTRL is adaptive in that it learns shared predictive structure between locations without requiring prior knowledge of dataset similarity. This is valuable in applied contexts where similarities may arise from latent or unobserved factors—such as administrative systems or economic patterns—and are difficult to encode directly. Additionally, CTRL is architecture-agnostic, making it a strong candidate for deployment in practical settings where constraints around cost or interpretability may affect what models are used.

We now examine some of the limitations of our approach. First while the runtime was reasonable for our applications, it may be prohibitive in settings with thousands of groups (See Section \ref{sec:runtime} for options to speed up).  Our out-of-distribution theory establishes that CTRL is robust to random distribution shifts, but does not extend to adversarial distribution shifts between locations.

To conclude, our work is motivated by a high-impact, real-world challenge: the geographic assignment of asylum seekers in Switzerland. We introduce CTRL to make progress on this problem and show it  maintains strong predictive accuracy while improving a key ranking metric, which is especially important for informing downstream decision-making tasks.\footnote{We aim to provide predictions that support—not replace—the work of our partners making critical downstream decisions. Any output from our model must be interpreted and finalized through the lens of stakeholder expertise.}
Lastly, to support further research and adaptation, we release our codebase along with adapted versions of four out of five datasets\footnote{The Swiss Asylum Seekers dataset is not publicly available, but can be requested from the Swiss State Secretariat for Migration.}: \nolinkurl{https://github.com/Gjain234/CtrlYourShift}.

\bibliographystyle{ACM-Reference-Format}
\bibliography{sample-base}

\appendix
\onecolumn

\section{Proof of Proposition~\ref{prop:leafwise-ctrl-risk}}\label{app:proof-leafwise}

\begin{proof}
Fix $w$ and target source $g$. By definition of $R^g$,
\begin{equation*}
Y-\hat f_{\mathrm{CTRL}}^w(X,g)
=Y-\hat f_{\text{base}}(X,g)-\hat f_{\text{residual}}^w(X)
=R^g-\hat f_{\text{residual}}^w(X).
\end{equation*}
Hence
\begin{equation*}
E_g\!\left[(Y-\hat f_{\mathrm{CTRL}}^w(X,g))^2\right]
=E_g\!\left[(R^g-\hat f_{\text{residual}}^w(X))^2\right].
\end{equation*}
Because functions in $\mathcal F$ are leafwise constant, write any $f\in\mathcal F$ as
\begin{equation*}
f(x)=\sum_{L\in\mathcal L} c_L\,\mathbf 1\{x\in L\},
\end{equation*}
for leaf values $\{c_L\}$. Then
\begin{equation*}
\sum_m w_m\hat E_m\!\left[(R^m-f(X))^2\right]
=\sum_{L\in\mathcal L}\sum_m w_m\hat P_m(L)
\Big(c_L-\hat\mu_{R,m}(L)\Big)^2 + C,
\end{equation*}
where
\begin{equation*}
\hat\mu_{R,m}(L):=\hat E_m[R^m\mid X\in L],
\end{equation*}
and $C$ does not depend on $\{c_L\}$. Hence the minimization decouples leaf-by-leaf, and for each $L$ we solve
\begin{equation*}
\min_{c\in\mathbb R}\;\sum_m a_{mL}(c-b_{mL})^2,
\qquad a_{mL}:=w_m\hat P_m(L),\; b_{mL}:=\hat\mu_{R,m}(L).
\end{equation*}
The first-order condition gives
\begin{equation*}
\sum_m a_{mL}(c_L^w-b_{mL})=0
\quad\Longrightarrow\quad
c_L^w=\frac{\sum_m w_m\hat P_m(L)\hat\mu_{R,m}(L)}{\sum_m w_m\hat P_m(L)}.
\end{equation*}
Now fix a source $m$. Writing any $u\in\mathcal F$ on leaf $L$ as $u(x)=d_L$ for $x\in L$, the source-specific problem is
\begin{equation*}
\min_{u\in\mathcal F}\,\hat E_m\!\left[(R^m-u(X))^2\right]
\iff
\min_{d_L\in\mathbb R}\,\hat P_m(L)\big(d_L-\hat\mu_{R,m}(L)\big)^2 + C_m,
\end{equation*}
so the minimizer is $d_L^*=\hat\mu_{R,m}(L)$. Hence, for $x\in L$,
\begin{equation*}
r^m(x)=\hat\mu_{R,m}(L).
\end{equation*}
Combining this with the expression for $c_L^w$ above,
\begin{equation*}
\hat f_{\text{residual}}^w(x)=c_L^w
=\frac{\sum_m w_m\hat P_m(L)\,r^m(x)}{\sum_m w_m\hat P_m(L)},
\qquad x\in L.
\end{equation*}
Define leaf-renormalized weights
\begin{equation*}
\tilde w_{mL}:=\frac{w_m\hat P_m(L)}{\sum_j w_j\hat P_j(L)}.
\end{equation*}
Then $\sum_m \tilde w_{mL}=1$ and
\begin{equation*}
\hat f_{\text{residual}}^w(x)=\sum_m \tilde w_{mL} r^m(x),\qquad x\in L.
\end{equation*}
Substituting this identity into the target-risk display gives
\begin{equation*}
E_g\!\left[(Y-\hat f_{\mathrm{CTRL}}^w(X,g))^2\right]
=E_g\!\left[\left(R^g-\sum_m \tilde w_{m,L(X)}\, r^m(X)\right)^2\right].
\end{equation*}
This proves the exact equality with leaf-renormalized weights.

For the asymptotic simplification, no covariate shift implies
\begin{equation*}
\hat P_m(L)=P_1(L)+o_p(1),
\qquad
\sum_j w_j\hat P_j(L)=P_1(L)+o_p(1),
\end{equation*}
so $\tilde w_{mL}=w_m+o_p(1)$ uniformly in $(m,L)$. Since $|Y|\le C_Y$ and all predictors above are leafwise averages of bounded outcomes, there is $C<\infty$ with $\sup_{m,x}|r^m(x)|\le C$. Therefore
\begin{equation*}
\left|\sum_m (\tilde w_{m,L(X)}-w_m)r^m(X)\right|
\le C\sum_m\left|\tilde w_{m,L(X)}-w_m\right|=o_p(1),
\end{equation*}
and expanding the squared loss yields
\begin{equation*}
E_g\!\left[\left(R^g-\sum_m \tilde w_{m,L(X)}r^m(X)\right)^2\right]
=
E_g\!\left[\left(R^g-\sum_m w_m r^m(X)\right)^2\right]+o_p(1).
\end{equation*}
\end{proof}

\section{Excess risk under distribution shift}\label{sec:excess-risk}

To theoretically characterize the tradeoff between dataset size and distribution shift in our objective~\eqref{eq:weights_generation}, we build on the random distributional shift model \citep{bansak2024learning,jeong2024out}. This is a stylized model that assumes that the likelihood ratio between the source distributions and target distributions is random. In brief, pooling increases the effective sample size but can introduce estimation error due to distributional mismatches between source and target locations. Our results show that the excess risk scales with (1) model complexity, (2) outcome noise, (3) the strength of distributional shift, and (4) the proportion of data drawn from shifted sources. Throughout this section, we use $\mathrm{Var}_g(\cdot)$ to denote variance under the target distribution $P_g$.

We will now list the main assumptions underlying a random distribution shift. 
For simplicity, we consider the target distribution $P_g$ as fixed and we have access to several randomly shifted source distributions $P_m$, $m=1,\ldots,|\mathcal{M}|$. 

\paragraph{Random distribution shift.}
Let $I_1,\ldots,I_K$ be a disjoint partitioning of the sample space $\mathcal{X} \times \mathcal{Y}$, with $P_g((X,Y) \in I_k) = 1/K$. We assume that as $K \rightarrow \infty$, the partitioning approximates all square integrable functions \footnote{This can be achieved by a standard partitioning approach. For instance, if $Y$ is a continuous random variable, endpoints of intervals $I_k$ can be defined via the $\frac{k-1}{K}$-th and $\frac{k}{K}$-th quantiles of $Y$.}, i.e. $\lim_{K \rightarrow \infty} E_g[(f(X,Y) - E[f(X,Y)|I_\bullet])^2] \rightarrow 0$ for all $f(X,Y) \in L^2(P_g)$.
For $(x,y) \in I_k$ let
\begin{equation*}
     P_m(x,y) = \frac{W_k^m}{\frac{1}{K} \sum_{k'=1}^K W_{k'}^m} P_g(x,y),
\end{equation*}
for some set of positive random weights $W_k^m$. We assume that the vectors $(W_k^1,\ldots, W_k^{|\mathcal{M}|})$ are i.i.d.\ across $k=1,\ldots,K$. We draw $n_m$ many $(X,Y)$ observations from $P_m$ for $m=1,\ldots,|\mathcal{M}|$. That is, conditional on $P_m$, the observations from location $m$ are independent.

We consider the simplified setting where each model computes a simple average of outcomes within a fixed leaf $L \in \mathcal{L}$, where $\mathcal{L}$ is a partition of $\mathcal{X}$ (e.g., as in regression trees with fixed splits). In practice, the fixed leaf assumption can be satisfied via sample splitting: tree structure is learned on one half of the data and leaf averages are estimated on the other, ensuring the splits are independent of the noise.
Let $\mathcal{C} \subseteq \mathcal{M}$ denote a fixed cluster and define the excess risk
\begin{equation*}
  \mathcal{E}_g = E_g \left[ (R^g - \frac{\sum_{m=1}^{|\mathcal{M}|} z_mr_{m}n_m}{\sum_{m=1}^{|\mathcal{M}|} z_{m}n_m})^2 \right] - E_g[( Y -  E_g[Y|\mathcal{L}]  )^2 ].
\end{equation*}
This corresponds to the difference of the expected value of our objective \eqref{eq:weights_generation} and the MSE of prediction $E_g[Y|\mathcal{L}]$, which is the optimal prediction for a tree with splits $\mathcal{L}$. The full proof can be found in Appendix \ref{full_proof}.

\begin{proposition}[Excess risk under distribution shift]\label{prop:excess-risk} Let $K, n_m \rightarrow \infty$ with $K/n_m \rightarrow c_m \in (0,\infty)$. Write $n_m^* = \lim n_m / \sum_{m'} n_{m'}$. Let $\text{Var}_g(Y) \in (0,\infty)$ and $P_g(X \in L ) > 0$ for all $L \in \mathcal{L}$. Then, $K \cdot \mathcal{E}_g$ converges to a non-degenerate random variable with mean
\begin{equation*}
\begin{aligned}
\left( \beta(\mathcal{C})^\intercal \Sigma^W \beta(\mathcal{C}) 
+ \sum_m \beta(\mathcal{C})_m^2 c_m \right) \cdot
\sum_{L \in \mathcal{L}} \text{Var}_g(Y \mid X \in L),
\end{aligned}
\end{equation*}
where $\beta(\mathcal{C})_m = \frac{1_{m \in \mathcal{C}} n_m^*}{ \sum_{m' \in \mathcal{C}} n_{m'}^*}$ and $\Sigma_{m,m'}^W = \text{Cov}(W_k^m,W_k^{m'})$.
\end{proposition}

For simplicity, let us consider the case where we have some data from the target $P_g$ and some data from another location $P_m$ and define $\sigma_{\text{avg}}^2 = \frac{1}{|\mathcal{L}|} \sum_{L \in \mathcal{L}} \text{Var}_g(Y| X \in L)$ as the average noise (over the leaves). In that case, the expected excess risk of training only on $g$'s data is $\frac{1}{n_g} \cdot | \mathcal{L} | \cdot \sigma_{\text{avg}}^2$.
If we train on data from $P_g$ and $P_m$, the expected excess risk is
\begin{align*}
  &\Bigg( \underbrace{\frac{\Sigma^W_{mm}}{K}}_{\text{shift strength}} \cdot 
  \underbrace{\frac{n_m^2}{(n_m + n_g)^2}}_{\text{prop. shifted data}} +  
  \underbrace{\frac{1}{n_g + n_m}}_{\text{pooled sam. size}} \Bigg) 
  \cdot \underbrace{| \mathcal{L} |}_{\text{\# params}} 
  \cdot \underbrace{\sigma_{\text{avg}}^2}_{\text{noise}}
\end{align*}
We can think about  $\frac{\Sigma^W_{mm}}{K}$ as the strength of distribution shift. If $n_m$ is much larger than $n_g$, then the distribution shift hurts performance more since the data from $P_m$ is a larger share of the total training data. Thus, there is a tradeoff between data quantity and data quality (measured via the strength of distribution shift between source $m$ and target $g$).

\section{Proof of Proposition~\ref{prop:excess-risk}}\label{full_proof}

\begin{proof}

For $X \in L$ we can decompose
\begin{align*}
    R^g &= Y 
    - \hat{E}_{\text{pooled}}[Y \mid X \in L] \\
        &= \left( Y 
        - E_g[Y \mid X \in L] \right) 
        + \left( E_g[Y \mid X \in L] \right. \\
        &\qquad \left. 
        - \hat{E}_{\text{pooled}}[Y \mid X \in L] \right),
\end{align*}
where $\hat{E}_{\text{pooled}}$ denotes the empirical mean over the pooled data.
Furthermore, for $X \in L$,
\begin{equation*}
    r_{m} = \hat E_{m}[Y| X \in L]  - \hat E_{\text{pooled}}[Y| X \in L]
\end{equation*}
Now, leveraging the CLT in \citet{qualityorquantity}, we can use a Taylor expansion 
\begin{align*}
    &E_g[ Y | X \in L ] - \hat E_{\text{pooled}}[ Y | X \in L ] \\
    &= \frac{E_g[ Y 1_{X \in L} ]}{E_g[1_{X \in L}]} - \frac{\hat{E}_{\text{pooled}}[ Y 1_{X \in L} ]}{\hat{E}_{\text{pooled}}[1_{X \in L}]} \\
&= 
\frac{
E_g[Y \cdot 1_{X \in L}] 
- \hat{E}_{\text{pooled}}[Y \cdot 1_{X \in L}]
}{
E_g[1_{X \in L}]
} \\
&\quad 
- \frac{
\left( E_g[1_{X \in L}] - \hat{E}_{\text{pooled}}[1_{X \in L}] \right) 
\cdot E_g[Y \cdot 1_{X \in L}]
}{
E_g[1_{X \in L}]^2
}
\\ &\quad + o_P(K^{-1/2}) \\
&= \left( E_g[ \phi_L] - \hat E_{\text{pooled}}[\phi_L]   \right) + o_P(K^{-1/2})
\end{align*}
Here, $\phi_L = \frac{1_{X \in L}}{P_g(X \in L)} ( Y - E_g[Y | X \in L]  )$. 

Analogously, for $X \in L$,
\begin{equation*}
    r_{m} = \hat E_m[\phi_L] - \hat E_{\text{pooled}}[\phi_L] + o_P(K^{-1/2}).
\end{equation*}
Thus,
\begin{align*}
    &\mathcal{E}_g(\mathcal{C}) \\
    &=  \sum_{L \in \mathcal{L}} P_g(X \in L) \cdot \Big( 
E_g[Y| X \in L ] - \hat{E}_\text{pooled}[Y|X \in L] \\
& \qquad - \sum_{m=1}^{|\mathcal{M}|}   \frac{z_m n_m}{\sum_{m'} n_{m'} z_{m'}} ( \hat{E}_{m}[Y| X \in L] -  \hat{E}_{\text{pooled}}[Y| X \in L]) \Big)^2 \\
&= \sum_{L \in \mathcal{L}} P_g(X \in L) \cdot \Big( 
E_g[\phi_L] - \hat{E}_{\text{pooled}}[\phi_L] \\ 
& \qquad - \sum_{m=1}^{|\mathcal{M}|} \frac{z_m n_m}{\sum_{m'} n_{m'} z_{m'}} \cdot (\hat{E}_m[\phi_L] - \hat{E}_\text{pooled}[\phi_L]) 
\Big)^2 \\
     &= \sum_{L \in \mathcal{L}} P_g(X \in L) \cdot \Big( 
E_g[\phi_L] 
- \sum_{m=1}^{|\mathcal{M}|} \frac{z_m n_m}{\sum_{m'} n_{m'} z_{m'}} \cdot \hat{E}_m[\phi_L] 
\Big)^2 \\
&\quad + o_P(K^{-1})
\\
\end{align*}
Now we can use a distributional CLT from \citep{qualityorquantity} to obtain that the vector
\begin{equation*}
 \left(   \sqrt{K} \left( E_g[\phi_L]  - \sum_{m=1}^{|\mathcal{M}|} \frac{z_m n_m}{\sum_{m'} n_{m'} z_{m'}}  \hat E_m[\phi_L]   \right) \right)_{L \in \mathcal{L}}
\end{equation*}
converge jointly (jointly across $L \in \mathcal{L}$) to a Gaussian random vector with componentwise variance
\begin{equation*}
(\beta(\mathcal{C})^\intercal \Sigma^W \beta(\mathcal{C}) + \sum_m \beta_m^2(\mathcal{C}) c_m ) \text{Var}_g(\phi_L).
\end{equation*}
Thus,
\begin{align*}
    &    K \left( E_g[\phi_L]  - \sum_{m=1}^{|\mathcal{M}|}  \frac{z_m n_m}{\sum_{m'} n_{m'} z_{m'}}  \hat E_m[\phi_L]   \right)^2  \\
    &\stackrel{d}{\rightarrow} \chi^2_1 \left((\beta(\mathcal{C})^\intercal \Sigma^W \beta(\mathcal{C}) + \sum_m \beta_m^2(\mathcal{C}) c_m ) \text{Var}_g(\phi_L) \right)
\end{align*}
Hence, 
\begin{align*}
        &K \sum_{L \in \mathcal{L}} P_g(X \in L)   \left( E_g[\phi_L]  - \sum_{m=1}^{|\mathcal{M}|}  \frac{z_m n_m}{\sum_{m'} n_{m'} z_{m'}} \hat E_m[\phi_L]   \right)^2 
\end{align*}
converges to a random variable with mean
\begin{equation*}
    \sum_{L \in \mathcal{L}} P_g(X \in L) (\beta(\mathcal{C})^\intercal \Sigma^W \beta(\mathcal{C}) + \sum_m \beta_m^2(\mathcal{C}) c_m ) \text{Var}_g(\phi_L).
\end{equation*}
Using that $P_g(X \in L) \text{Var}_g(\phi_L) = \text{Var}_g(Y| X \in L) $, we can rewrite the previous display as
\begin{equation*}
    \sum_{L \in \mathcal{L}}  (\beta(\mathcal{C})^\intercal \Sigma^W \beta(\mathcal{C}) + \sum_m \beta_m^2(\mathcal{C}) c_m ) \text{Var}_g(Y | X \in L).
\end{equation*}
This completes the proof.
\end{proof}

\section{RWA across different thresholds}
In the main paper, we report RWA results using the top 20\% of individuals per location. This threshold provides a practical balance: it captures meaningful location-specific variation in predicted outcomes while avoiding the inclusion of too many individuals, which could obscure differences by averaging them out. To examine the sensitivity of our findings to this choice, we also report RWA results for a range of threshold levels in Table~\ref{tab:combined-rwa-all-clean}. Across all thresholds, CTRL either beats or closely matches all comparison methods.

\begin{table*}[ht!]
\centering
\caption{Average RWA across different top percentages for all datasets. Bold indicates best performance within each dataset and column.}
\renewcommand{\arraystretch}{1.0}
\setlength{\tabcolsep}{2pt}
\resizebox{\textwidth}{!}{
\begin{tabular}{l|cccc|cccc|cccc|cccc|cccc}
\hline
\textbf{Model} &
\multicolumn{4}{c|}{\textbf{Top 10\%}} &
\multicolumn{4}{c|}{\textbf{Top 20\%}} &
\multicolumn{4}{c|}{\textbf{Top 30\%}} &
\multicolumn{4}{c|}{\textbf{Top 40\%}} &
\multicolumn{4}{c}{\textbf{Top 50\%}} \\
& Reg & Tree & RF & BART & Reg & Tree & RF & BART &
  Reg & Tree & RF & BART & Reg & Tree & RF & BART &
  Reg & Tree & RF & BART \\
\hline
\multicolumn{21}{c}{\textbf{Synthetic Dataset}} \\
JTT & 0.819 & 0.622 & 0.818 & 0.836 & 0.783 & 0.618 & 0.777 & 0.798 & 0.750 & 0.611 & 0.741 & 0.760 & 0.714 & 0.602 & 0.707 & 0.723 & 0.678 & 0.585 & 0.672 & 0.685 \\
RWG & 0.819 & 0.693 & 0.803 & 0.820 & 0.783 & 0.683 & 0.765 & 0.775 & 0.750 & 0.665 & 0.731 & 0.736 & 0.714 & 0.648 & 0.697 & 0.700 & 0.678 & 0.626 & 0.663 & 0.666 \\
Global & 0.820 & 0.724 & 0.805 & 0.863 & 0.783 & 0.705 & 0.770 & 0.819 & 0.750 & 0.681 & 0.737 & 0.778 & 0.714 & 0.664 & 0.704 & 0.738 & 0.678 & 0.641 & 0.670 & 0.698 \\
Local & 0.984 & 0.676 & \textbf{0.954} & 0.967 & 0.958 & 0.690 & 0.912 & 0.926 & 0.918 & 0.689 & 0.865 & 0.880 & 0.870 & 0.672 & 0.815 & 0.830 & 0.811 & 0.646 & 0.764 & 0.777 \\
TRL & 0.984 & 0.777 & 0.938 & 0.968 & 0.958 & 0.746 & 0.896 & 0.929 & 0.919 & 0.717 & 0.853 & 0.883 & 0.871 & 0.686 & 0.805 & 0.831 & 0.810 & 0.656 & 0.754 & 0.777 \\
CTRL & \textbf{0.987} & \textbf{0.812} & 0.952 & \textbf{0.975} & \textbf{0.962} & \textbf{0.772} & \textbf{0.913} & \textbf{0.939} & \textbf{0.922} & \textbf{0.736} & \textbf{0.870} & \textbf{0.893} & \textbf{0.873} & \textbf{0.701} & \textbf{0.821} & \textbf{0.841} & \textbf{0.813} & \textbf{0.666} & \textbf{0.767} & \textbf{0.785} \\
\hline
\multicolumn{21}{c}{\textbf{Swiss Asylum Seekers Dataset}} \\
JTT & 0.357 & 0.220 & 0.493 & 0.450 & 0.288 & 0.230 & 0.391 & 0.363 & 0.257 & 0.239 & 0.327 & 0.311 & 0.237 & 0.238 & 0.282 & 0.273 & 0.216 & 0.223 & 0.246 & 0.242 \\
RWG & 0.356 & 0.392 & 0.486 & 0.443 & 0.296 & 0.322 & 0.384 & 0.362 & 0.264 & 0.285 & 0.325 & 0.309 & 0.242 & 0.242 & 0.283 & 0.271 & 0.220 & 0.212 & 0.247 & 0.239 \\
Global & 0.356 & 0.397 & 0.495 & 0.448 & 0.295 & 0.331 & 0.394 & 0.367 & 0.264 & 0.284 & 0.332 & 0.312 & 0.241 & 0.252 & 0.284 & 0.273 & 0.219 & 0.227 & 0.249 & 0.241 \\
Local & 0.356 & 0.381 & 0.469 & 0.425 & 0.297 & 0.306 & 0.375 & 0.346 & 0.264 & 0.269 & 0.322 & 0.299 & 0.239 & 0.239 & 0.280 & 0.265 & 0.217 & 0.215 & 0.247 & 0.236 \\
TRL & \textbf{0.376} & 0.411 & 0.500 & \textbf{0.461} & 0.310 & 0.345 & 0.400 & \textbf{0.373} & 0.275 & 0.298 & \textbf{0.338} & \textbf{0.319} & \textbf{0.247} & 0.261 & \textbf{0.290} & 0.276 & \textbf{0.224} & 0.232 & \textbf{0.251} & \textbf{0.243} \\
CTRL & \textbf{0.376} & \textbf{0.420} & \textbf{0.503} & 0.455 & \textbf{0.312} & \textbf{0.348} & \textbf{0.401} & 0.372 & \textbf{0.276} & \textbf{0.299} & 0.337 & 0.318 & \textbf{0.247} & \textbf{0.262} & 0.289 & \textbf{0.277} & \textbf{0.224} & \textbf{0.233} & \textbf{0.251} & 0.242 \\
\hline
\multicolumn{21}{c}{\textbf{Education Dataset}} \\
JTT & \textbf{0.955} & \textbf{0.938} & 0.920 & \textbf{0.964} & 0.937 & 0.916 & 0.907 & \textbf{0.946} & 0.918 & 0.877 & 0.893 & 0.927 & 0.902 & 0.853 & 0.875 & 0.908 & 0.887 & 0.831 & 0.856 & 0.890 \\
RWG & \textbf{0.955} & 0.891 & 0.963 & 0.960 & 0.939 & 0.885 & 0.941 & 0.941 & 0.918 & 0.877 & 0.921 & 0.920 & 0.903 & 0.860 & 0.902 & 0.903 & 0.887 & 0.855 & 0.888 & 0.888 \\
Global & \textbf{0.955} & 0.856 & \textbf{0.966} & 0.961 & 0.940 & 0.854 & 0.943 & 0.944 & 0.918 & 0.852 & 0.926 & 0.925 & 0.903 & 0.844 & 0.908 & 0.906 & 0.887 & 0.842 & \textbf{0.891} & 0.889 \\
Local & 0.901 & 0.879 & 0.944 & 0.954 & 0.898 & 0.875 & 0.931 & 0.942 & 0.889 & 0.873 & 0.916 & 0.928 & 0.875 & 0.856 & 0.896 & 0.910 & 0.861 & 0.837 & 0.883 & 0.892 \\
TRL & 0.953 & 0.934 & 0.964 & 0.956 & \textbf{0.941} & 0.927 & 0.945 & 0.944 & 0.925 & 0.913 & 0.928 & 0.929 & 0.906 & 0.890 & \textbf{0.909} & \textbf{0.911} & \textbf{0.890} & 0.874 & \textbf{0.891} & \textbf{0.893} \\
CTRL & 0.952 & 0.937 & 0.965 & 0.956 & \textbf{0.941} & \textbf{0.931} & \textbf{0.946} & 0.944 & \textbf{0.926} & \textbf{0.918} & \textbf{0.929} & \textbf{0.930} & \textbf{0.907} & \textbf{0.895} & \textbf{0.909} & \textbf{0.911} & \textbf{0.890} & \textbf{0.881} & \textbf{0.891} & \textbf{0.893} \\
\hline
\end{tabular}
}
\label{tab:combined-rwa-all-clean}
\end{table*}

\begin{table*}[t]
\centering
\caption{Overview of datasets used in our experiments. Precise numbers in the Swiss Asylum Seeker dataset are withheld due to data restrictions.}
\resizebox{\textwidth}{!}{\begin{tabular}{
  >{\raggedright\arraybackslash}p{2.8cm} |
  >{\raggedright\arraybackslash}p{3.2cm} |
  >{\raggedright\arraybackslash}p{3.2cm} |
  >{\raggedright\arraybackslash}p{2.2cm} |
  c | c | c |
  >{\raggedright\arraybackslash}p{3.2cm}
}
\toprule
\textbf{Title} & \textbf{Outcome} & \textbf{Sources} & \textbf{Source Info} & \textbf{Overall Size} & \textbf{Train Size} & \textbf{Test Size} & \textbf{Features} \\
\midrule
Synthetic & simulated outcome & simulated distributional shifts & 50 sources (sizes 40-2000) & 40,000 & 13,320 & 26,680 & 20 random continuous features \\
\midrule
Swiss Asylum Seekers & binary two year employment outcome & cantons & 26 cantons (sizes 50-3900) & $\sim$30,000 & $\sim$24,000 & $\sim$6000 & 135 continuous or one hot encoded variables - demographics, language, arrival time, administrative info, etc. \\
\midrule
Education & binary high school graduate status & US states (including PR) & 51 states/territories (sizes 244-122092) & 470,442 & 94,089 & 376,353 & 13 one hot encoded variables -  demographic, ethnicity, citizenship, language, disability \\
\midrule
UK Asylum Decisions & binary asylum approval outcome & nationality & 103 nationalities (sizes 41-1918) & 84,449 & 42,225 & 42,224 & 108 one hot encoded and continuous variables features - demographics, administrative info, arrival time \\
\midrule
Dissecting Health Bias & binary indicator of chronic illness & race+gender+age demographic group (e.g. "18-24\_0\_black" is a group for age 18-24 men that are black) & 28 groups (sizes 76-6453) & 47,865 & 28,719 & 19,146 & 93 one hot encoded and continuous variables - risk score, health indicators, enrolled program \\
\bottomrule
\end{tabular}
}
\label{tab:dataset-summary}
\end{table*}

At the 10\% threshold, the Global and JTT models perform comparatively well on the Education dataset. This is not unexpected, as the top 10\% of individuals are likely those who would achieve good outcomes in most locations, making globally optimized models like Global and JTT more effective in this narrow slice. Even in this case, CTRL performs similarly, highlighting its effectiveness across different evaluation settings. Lastly, these results are shown only for datasets that involve an explicit decision-making task: the synthetic dataset, the Swiss Asylum Seekers dataset, and the Education dataset.
\section{Datasets} We make four of the five datasets used in this paper available through our codebase: \url{https://github.com/Gjain234/CtrlYourShift}, along with a README containing instructions for accessing all datasets. Table~\ref{tab:dataset-summary} also provides a detailed overview of each dataset, including information on features, outcomes, source structure, and sample size. The original Education, Dissecting Bias in Health, and UK Asylum Decisions datasets are publicly accessible, and links to their original sources can be found in the paper's citations and the codebase README. We additionally provide adapted versions of the datasets that reflect a multi-source structure, where each source has a unique relationship between its features and outcome variable. We hope this can help facilitate further research on prediction tasks for datasets with imbalanced and heterogeneous sources.

We selected these datasets to reflect both our real-world motivation and broader methodological relevance. The Swiss Asylum Seekers dataset directly motivated the development of CTRL, as this work emerged from a real-world collaboration focused on improving decision-making for asylum seeker placement in Switzerland, where data sparsity and heterogeneity across locations present significant challenges. The Education data is an adapted version of United States Census data for which we turn to \citet{ding2021retiring} for pre-processing. This dataset is commonly used for studying debiasing, fairness, and distribution shift, making it a natural testbed for us \cite{NEURIPS2023_a134eaeb,jeong2024out}. The Dissecting Bias in Health datasets was similarly constructed for this sort of analysis \cite{obermeyer}. Finally, we introduce the UK Asylum Decisions dataset, which, to the best of our knowledge, has not been used in prior machine learning research. However, it naturally exhibits both distribution shift and many-source structure, making it a promising and relevant benchmark for researchers interested in studying these challenges and exploring new real-world datasets. Its inclusion supports a more concrete observation: although CTRL was originally developed with a specific application in mind, many real-world datasets naturally take the form of imbalanced, unevenly sized many-source settings. This paper takes a step toward developing practical tools that are explicitly designed to handle such structure. 
Due to privacy constraints, we are unable to release the Swiss Asylum Seekers dataset, as it contains sensitive, non-public information.

\section{Synthetic Data Generation}\label{synthetic_details}
To evaluate CTRL in a more controlled setting, we construct a synthetic dataset designed to reflect the key challenge CTRL aims to address: learning accurate, location-specific predictions when data availability is highly uneven across sources. We generate $N = 40{,}000$ individual observations assigned to $|\mathcal{M}| = 50$ locations, each denoted by $g \in \mathcal{M}$, with a minimum location size of 120 and the max around 3000. Location sizes are drawn from a Pareto distribution to simulate naturally imbalanced settings.

Each individual is assigned a $d = 20$-dimensional feature vector drawn from a standard normal distribution. Outcome probabilities are computed as a weighted combination of global and location-specific signal components, passed through a sigmoid to map values to $[0,1]$, with binary outcomes sampled accordingly. A subset of locations are grouped into latent clusters of size 2–7, where each location $g$ shares a common base weight vector with small random shifts. The remaining locations are assigned weights independently, with each location having its own feature-level means and variances. For the experiments in this paper, we use a weighting of 0.3 on the global signal and 0.7 on the local (location-specific) signal when combining probabilities. We selected this ratio based on empirical observations that our baseline TRL achieved greater performance gains under this structure, suggesting that this weighting provides a favorable balance between pooled information and location-specific variation. It reflects a setting where shared signal is informative but insufficient on its own, requiring local adaptation to fully capture heterogeneous outcome patterns. 

While this dataset is constructed to represent the types of structure CTRL is designed to handle—namely, latent similarity across some locations—it is important to note that the data is not generated using the CTRL algorithm, nor does it rely on residual modeling. In fact, there is no explicit notion of residuals in the data-generating process. Because each location’s outcome is generated using a linear transformation of features, the true model for each group $g$ is just a linear regression. This remains true even when combining pooled and location-specific components—since the weighted sum of linear functions is itself linear. Therefore, with sufficient data per location, location-level linear models would be the optimal choice. This is reflected in our empirical results (Table~\ref{tab:all_mse_smallmse}), where local linear model did actually perform best in overall MSE. However, in small-sample regimes local models degrade rapidly, and CTRL's advantage becomes clear. By adaptively pooling data based on similarity across locations, CTRL is able to improve prediction performance—even in cases where the data was not explicitly generated to include residual-based or hierarchical structure—demonstrating its adaptability to realistic many-source (or \textit{many-location}) settings.

We include the generated dataset and the full script used to construct it, `generate\_synthetic\_data.py`, in our codebase. The script allows users to easily reproduce or modify the synthetic data for further experimentation.
\section{Simple Clustering Baselines} \label{simpleclustering}
As an additional benchmark, we implement two simple clustering baselines that group locations using standard notions of similarity between their residual distributions: one based on the Wasserstein distance and the other on a simple correlation measure. For both baselines, we first train a global model on the full training cohort and compute residuals as the difference between the observed outcomes and the corresponding global predictions.

For the Wasserstein variant, we treat the empirical residual distribution at each location as a summary of systematic model error and compute pairwise Wasserstein-1 distances between locations. For the correlation-based variant, we construct residual vectors for each location over the full dataset and compute pairwise correlations between these vectors.

Rather than explicitly clustering the resulting distance matrices, we directly evaluate the quality of these similarity measures by comparing the top-ranked source locations for each target against the ground-truth clusters in the synthetic data. This allows us to assess whether the induced distances meaningfully capture cluster structure without introducing additional clustering hyperparameters. Together, these baselines provide feature-agnostic approaches that pool locations solely based on residual similarity, without target-specific adaptation or optimization.
\section{Overall Performance}
\begin{figure}\centering
\includegraphics[width=.5\linewidth]{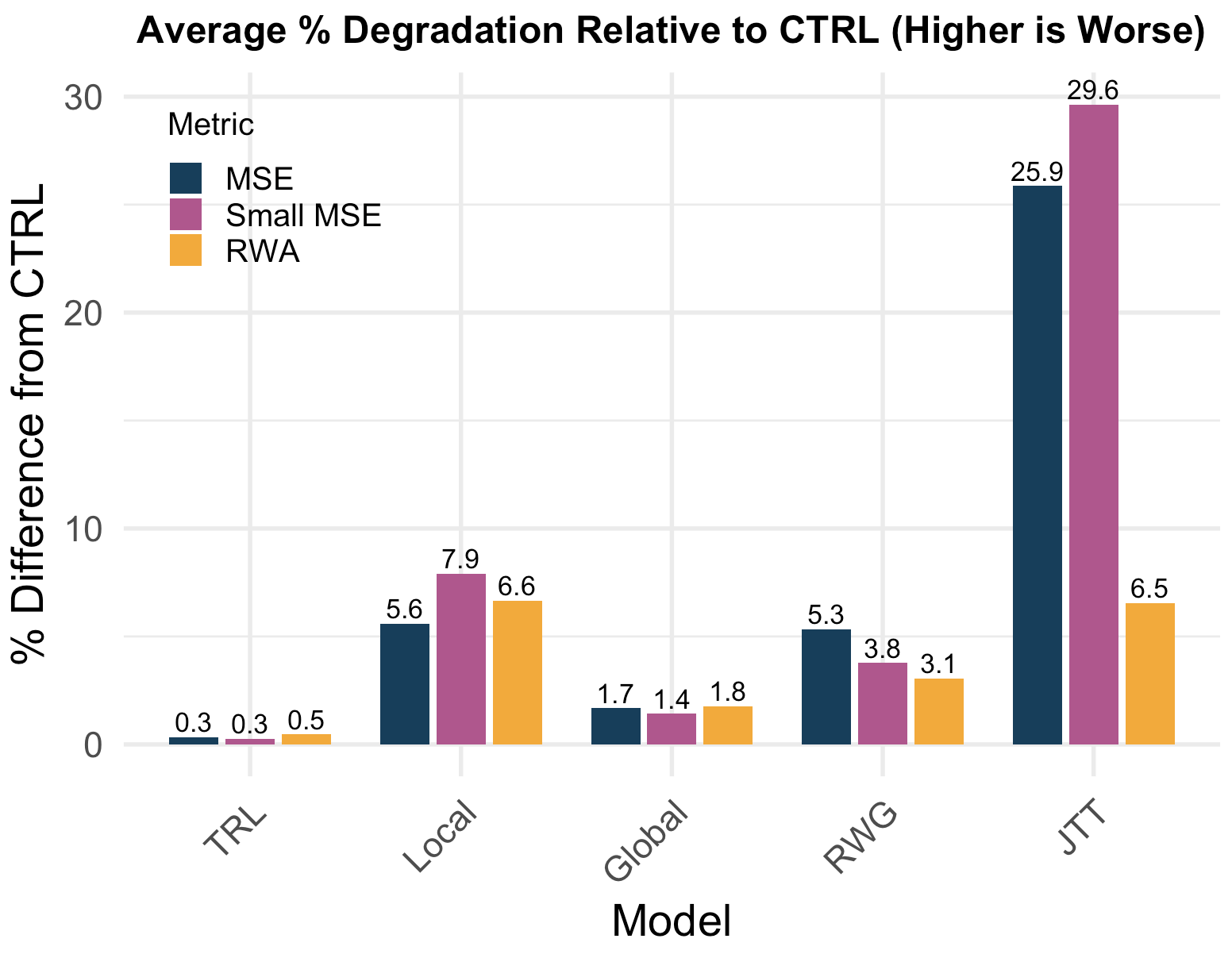}
  \caption{Average performance gaps  relative to CTRL are shown across all applicable datasets for each metric, excluding the synthetic dataset, which was specifically constructed for our use case. A positive value here indicates worse performance.}
  \label{fig:performance_gap}
\end{figure}

Evaluating CTRL across multiple datasets is essential for assessing its robustness and broader applicability. In addition to dataset-specific analyses, we summarize CTRL’s average performance across all datasets to provide a more holistic view. Figure~\ref{fig:avg_performance} presents an average rank metric, while Figure~\ref{fig:performance_gap} displays the average performance gap between CTRL and each benchmark method. We exclude the synthetic dataset from this comparison, as it was explicitly constructed to align with our modeling assumptions.

In Figure~\ref{fig:performance_gap}, positive values indicate that a benchmark performs worse than CTRL. Across all evaluation metrics—MSE, Small MSE, and RWA—all benchmarks underperform relative to CTRL on average. The improvement in the RWA metric is particularly noteworthy, as it directly reflects the model’s ability to rank people at different locations effectively. Given our goal of improving decision-making for asylum seeker placement in Switzerland, this result highlights CTRL’s potential to inform real-world allocation policies.

\section{Combinatorial Optimization} \label{appendix:combinatorial_optimization_details}
In order to solve Problem \ref{eq:weights_generation} we need to utilize the NonConvex solver with Gurobi. Additionally, running it for all groups ends up becoming extremely slow, so we use $\lambda=$5-7 groups for every run depending on how large the dataset is (more than this and the optimization problems takes much longer to solve). After repeating $\gamma$ times, we can average over all the runs to get an overall location ranking even though each individual run only looks at a subset.
\section{CTRL Parameters} \label{clustered_x_param}

\begin{figure}[ht!]
  \centering
\includegraphics[width=.5\linewidth]{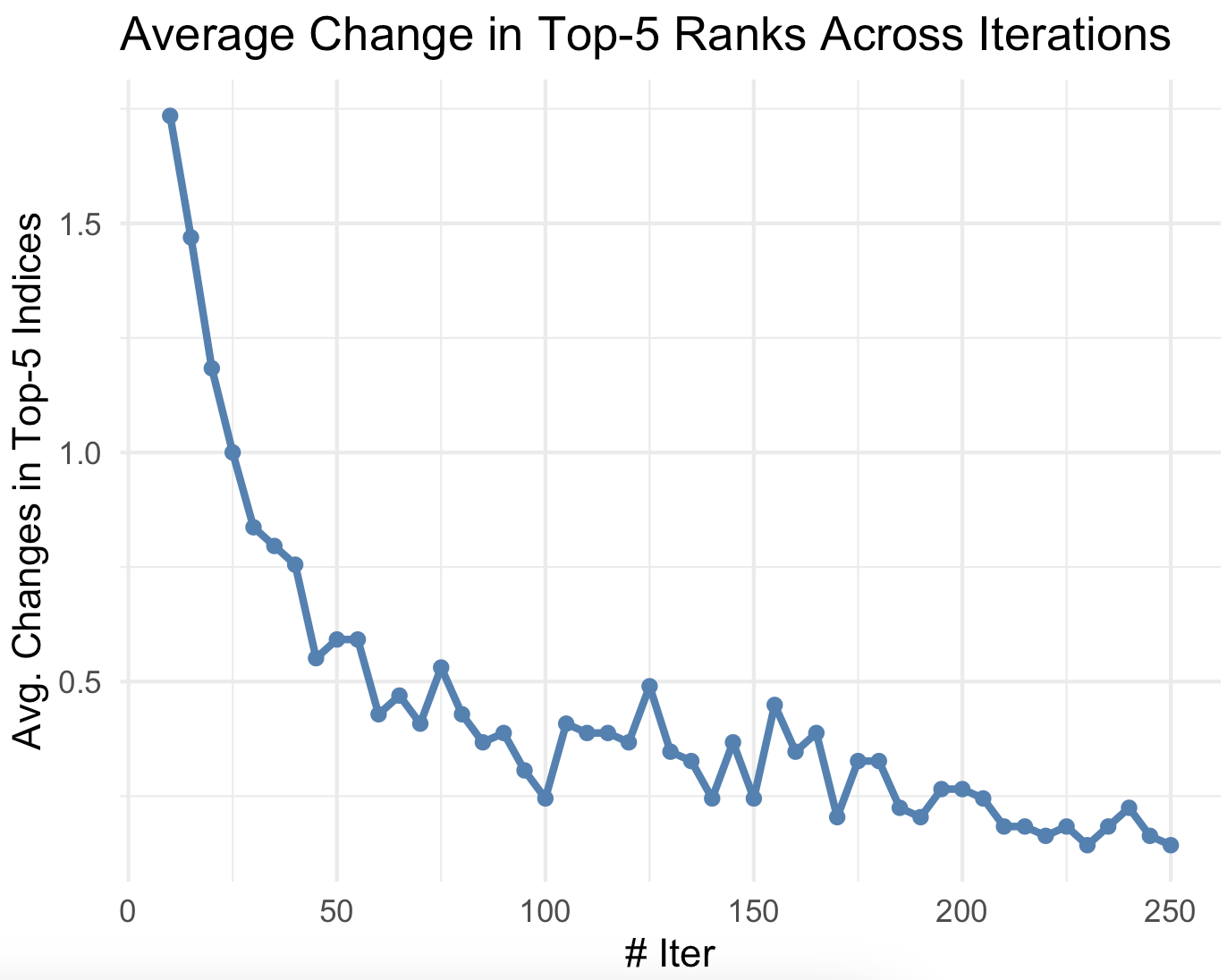}
  \caption{Change in top 5 locations picked as we add more iterations for running Algorithm \ref{alg:xlearn_clustering}. We reach about 0.25 changes per location after 150 iterations.}
  \label{fig:top_5_metric}
\end{figure}

We conduct simulations on synthetic data to guide the selection of optimal hyperparameters for our CTRL framework. For generating the initial weights $\mathbf{w_g}$, we rerun Algorithm~\ref{alg:xlearn_clustering} $\gamma=250$ times. This decision is motivated by empirical observations: after approximately 100 iterations, the average change in the top five source locations (as ranked for each target location’s cluster) stabilizes to around 0.25 locations (see Figure~\ref{fig:top_5_metric}). This suggests that beyond this point, the top-ranked source locations remain largely consistent. We use the average change in the top 5 as our stability metric because cluster sizes tend to average around five (see Figure \ref{fig:state_group} for cluster examples), making it more critical to identify the correct set of source locations than to determine their exact order within the ranking. Thus, our metric prioritizes inclusion over position.

\begin{figure}[ht!]
  \centering
\includegraphics[width=.5\linewidth]{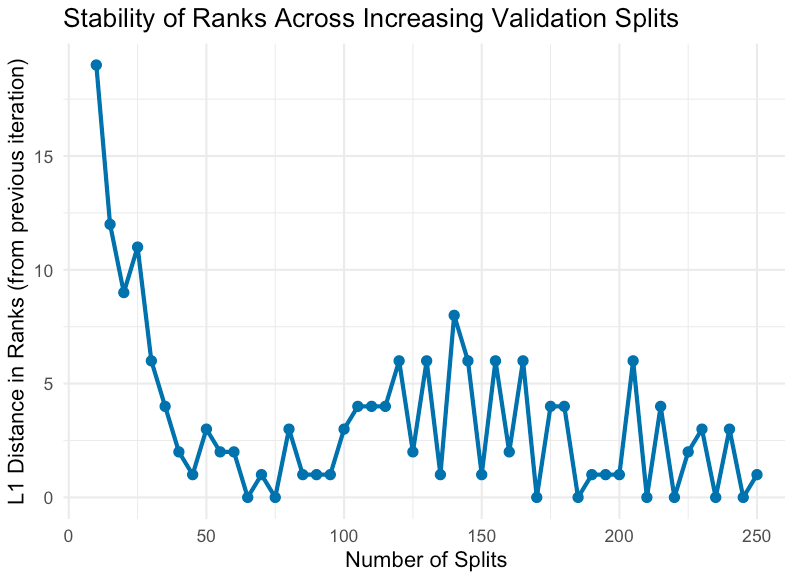}
  \caption{Change in ranks vector $\mathbf{k^*}$ returned by Algorithm \ref{alg:cluster_selection} for all locations $g \in \mathcal{M}$. L1 distance between $\mathbf{k^*}$ and $\mathbf{k^{*}}^{(+5)}$ corresponding to the $i$'th and $i+5$'th validation split.}
  \label{fig:rank_stability}
\end{figure}

For estimating the mean squared error (MSE) in Algorithm~\ref{alg:cluster_selection}, we also select 250 iterations. As shown in Figure~\ref{fig:rank_stability}, the L1 change in consecutive $\mathbf{k^*}$'s—the vector of optimal top-$k$ values ($k^*_\ell$) across all target locations—decreases and converges to under 5 after approximately 50 iterations. Given that there are 50 target locations in the synthetic dataset, this implies that the expected change in $k^*_\ell$ for a given location with each additional iteration is about 0.1. Therefore, we consider any iteration count beyond 50 to yield stable and reliable results, and choose 250 to match the splits we use for the $\mathbf{w_g}$ generation.

\begin{figure}[ht!]
  \centering
\includegraphics[width=.5\linewidth]{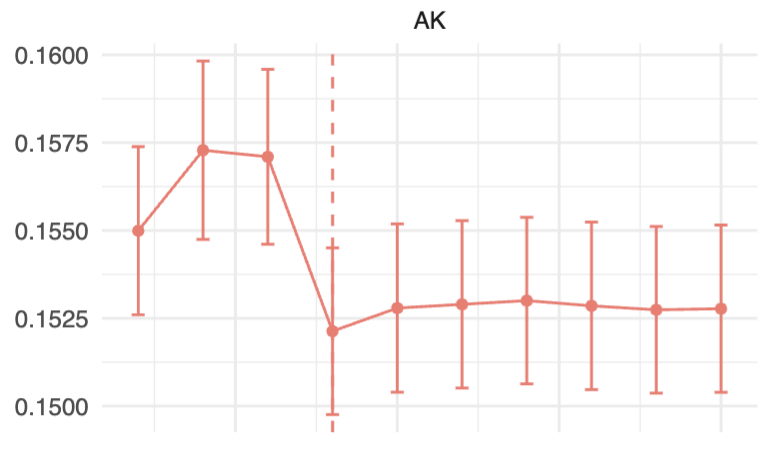}
  \caption{Change in cross validated MSE as we add more source locations to train $\hat{f}_\text{residual}^{\mathcal{C}(g)}$ for the state of Alaska. The dotted line represents the actual $k^*$ picked for Alaska.}
  \label{fig:top_k}
\end{figure}

We constrain the value of $K$ in Algorithm~\ref{alg:cluster_selection} to be at most 10. Empirically, we observe that beyond this threshold, the clusters begin to resemble those from a global model, diminishing the benefits of location-specific modeling. Therefore, to preserve the uniqueness of each target location, we limit the number of source locations considered for its cluster to 10. Figure~\ref{fig:top_k} illustrates this trend using the state of Alaska from the education dataset: the MSE plateaus after only five source locations are included, a pattern that holds across most target locations. Nonetheless, we allow $K$ to reach up to 10 to provide some buffer. In practice, we find that most location clusters are around or less than 5 source locations. For reference, all generated clusters are analyzed and visualized in the following section.


\section{Qualitative Analysis of Clusters} \label{qual_analysis}

\begin{figure*}\centering
  \includegraphics[width=1\textwidth]{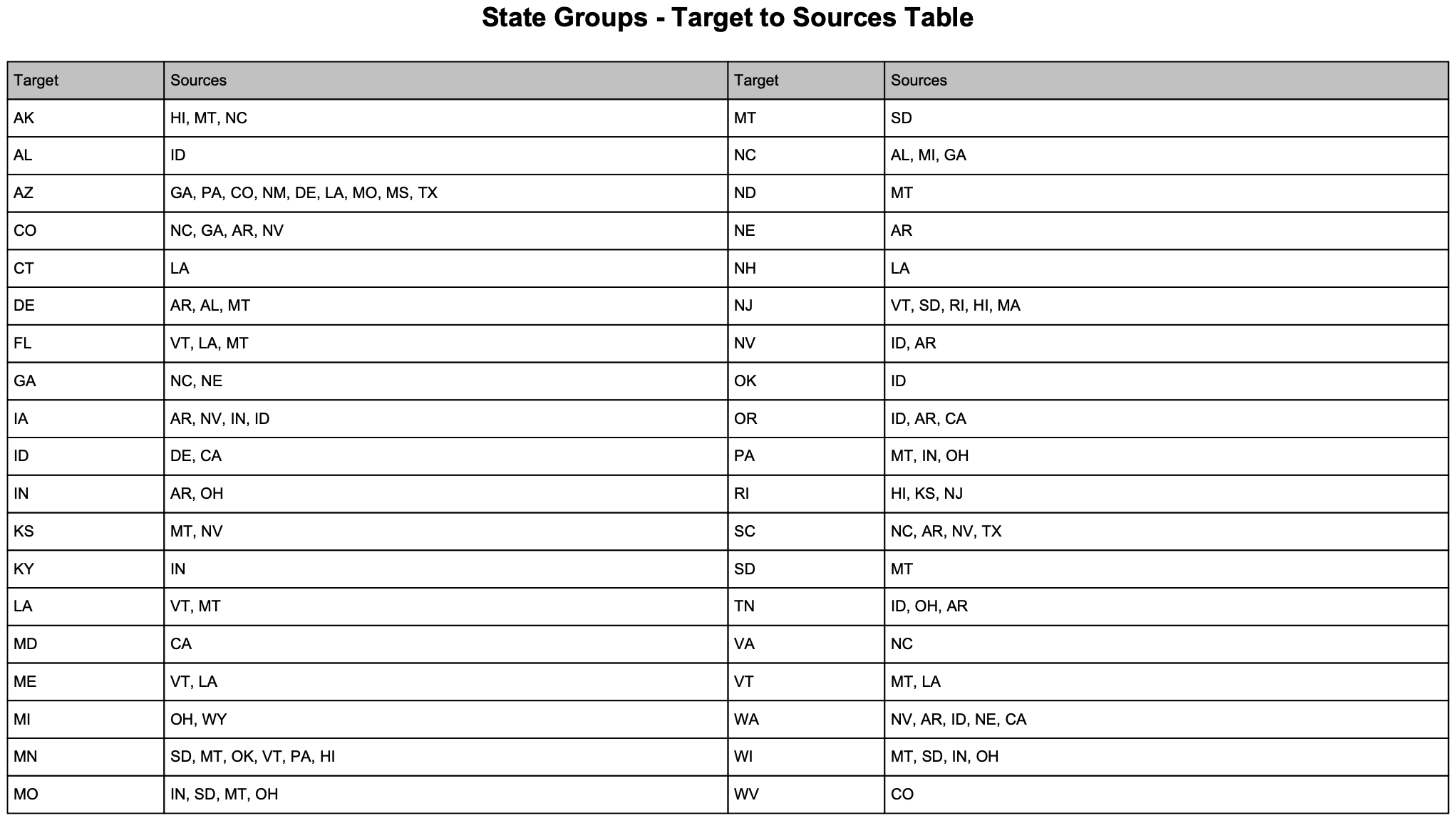}
  \caption{Table showing the source state clusters created by CTRL for each target state in the Education datasets. We omit states that only picked themselves for their cluster. Every state is required to pick itself as a source (see Equation \ref{eq:weights_generation}), but we don't explicitly write that in the table for simplicity. In demonstrating our method, we drop all sources that have too few rows (less than 40) to reliably provide predictions for and therefore those sources also do not exist in these tables.
}
  \label{fig:state_group}
\end{figure*}

\begin{figure*}\centering
  \includegraphics[width=0.4\textwidth]{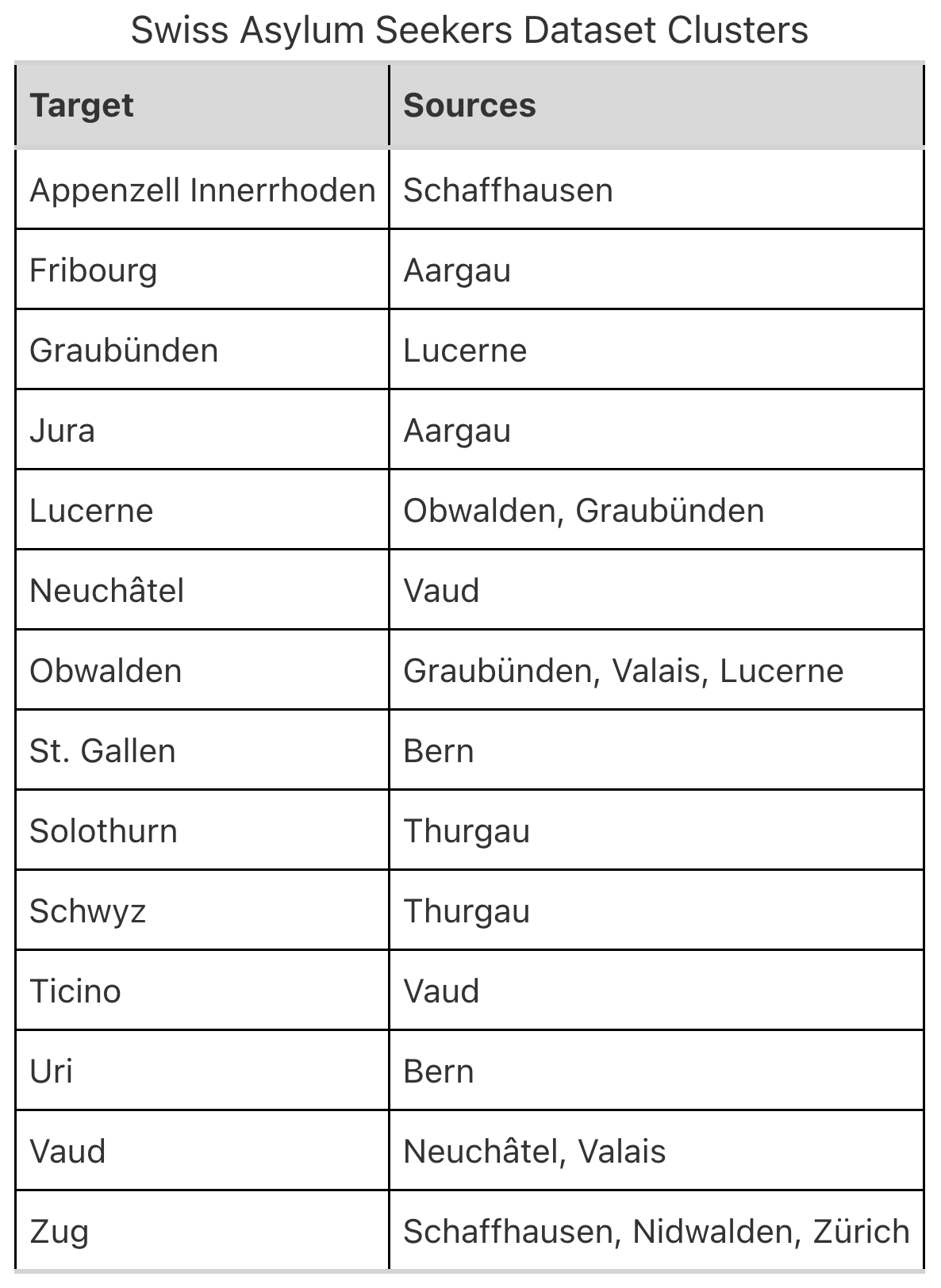}
  \caption{Table showing the canton clusters created by CTRL in the Swiss Asylum Seekers dataset. We again omit the actual target in the source set even though it is always used}
  \label{fig:swiss_group}
\end{figure*}

\begin{figure*}\centering
  \includegraphics[width=0.4\textwidth]{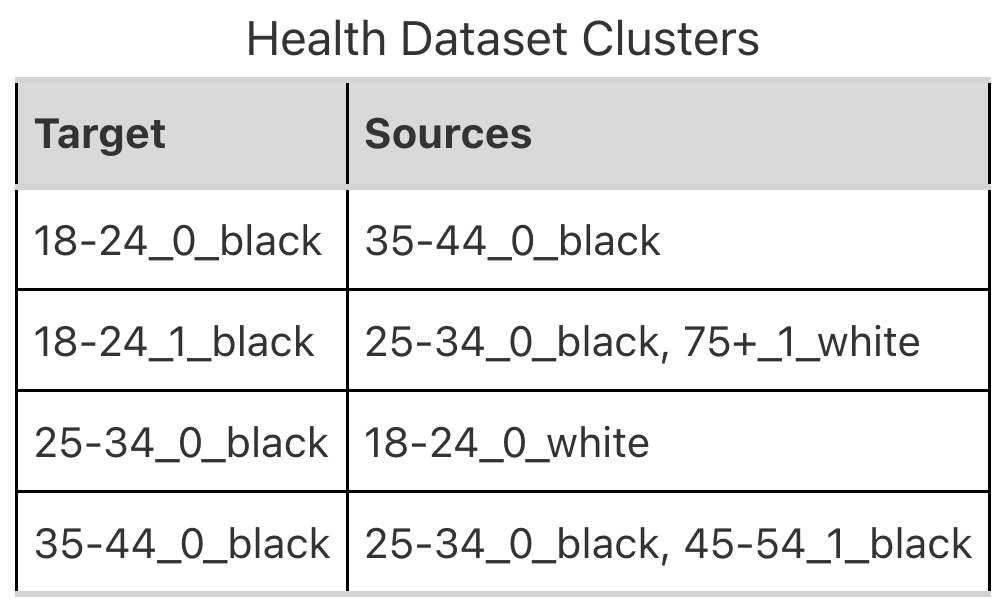}
  \caption{Table showing the group clusters created by CTRL in the Dissecting Health Bias dataset. 
The groups are formatted \textless age group\textgreater\_\textless 1 if woman\textgreater\_\textless race\textgreater.
}
  \label{fig:health_group}
\end{figure*}

\begin{figure*}\centering
  \includegraphics[width=1\textwidth]{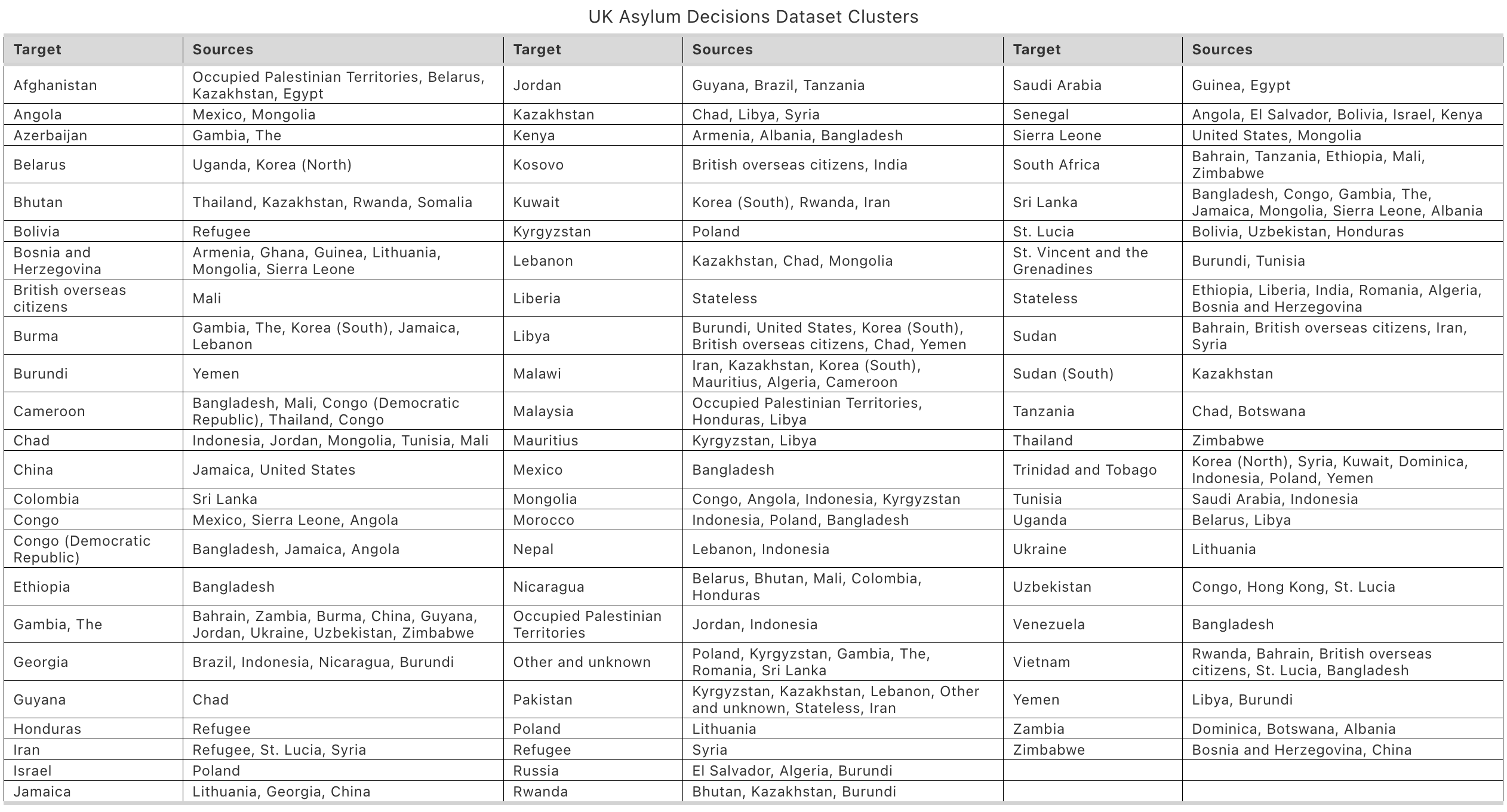}
  \caption{Table showing the nationality clusters created by CTRL in the UK Asylum Decisions dataset.}
  \label{fig:uk_group}
\end{figure*}

We provide the full set of cluster assignments produced by CTRL for each dataset in Figures~\ref{fig:state_group}, \ref{fig:swiss_group}, \ref{fig:health_group}, and \ref{fig:uk_group}. We don't provide this for the synthetic dataset as they do not have any qualitative meaning. These tables list, for each target, the set of source groups selected by CTRL (excluding the trivial case in which a target selects only itself). Additionally, although each target is always included in its own source set, we omit this label for clarity. Together, these tables offer a complete view of how CTRL allocates auxiliary data across locations or groups.

It is important to emphasize that these source sets should not be interpreted as reflecting geographic, demographic, or semantic similarity between locations or groups. CTRL forms clusters exclusively on the basis of complementary conditional prediction behavior—specifically, similarity in $P(Y \mid X)$ patterns—rather than any interpretable notion of proximity or shared attributes. As a result, the induced clusters are not intended to correspond to meaningful groupings in a substantive or domain-specific sense. Indeed, a central motivation for CTRL is the recognition that articulating human-interpretable criteria for when two units are “similar” is often ill-posed and practically infeasible in real-world policy settings. Consequently, a comprehensive qualitative interpretation of all clusters across all datasets is neither realistic nor conceptually well-defined.

Nevertheless, inspecting these clusters can still serve as a useful qualitative diagnostic, helping to illustrate how CTRL leverages auxiliary information and whether the resulting groupings exhibit any intuitive or unexpected structure. To this end, we present a more detailed qualitative examination of the \textbf{Education} dataset as a representative exercise. This setting allows us to assess, at a high level, whether the clustering behavior produced by CTRL aligns with plausible patterns, without over-interpreting the results.

In Figure~\ref{fig:state_group}, Alaska appears in the first row and selects, in order of importance, Hawaii, Montana, and North Carolina as its source states. While Alaska and Hawaii are geographically and culturally distinct, they share several structural characteristics, including geographic separation from the continental United States, relatively recent statehood, diverse populations, and overlapping economic sectors such as tourism and military presence. These commonalities may help explain why they align under a data-driven notion of conditional predictive similarity, despite clear differences along other dimensions.

Another notable example is Arizona, which selects a relatively large number of source states (including itself). This behavior may reflect the state’s demographic and economic diversity, allowing it to benefit from auxiliary information drawn from a wide range of locations. Minnesota exhibits a similar pattern. Some clusters display partial geographic coherence, though not consistently enough to justify a manually specified regional structure. For instance, Washington selects California, Nevada, Arkansas, Idaho, and Nebraska—of which several, but not all, are geographically proximate.

We also briefly note an interesting pattern in the \textbf{Dissecting Health Bias} dataset (Figure~\ref{fig:health_group}), where clustering occurs almost exclusively among groups identified as Black. This behavior is consistent with the structure of the dataset, which highlights lower data availability and quality for these groups. In this setting, CTRL appears to prioritize pooling auxiliary information for underrepresented populations, a behavior that may help improve predictive performance and mitigate disparities driven by data scarcity. 

While we do not attempt a comprehensive qualitative analysis of all clusters across all datasets, we include the complete set of cluster assignments produced by CTRL for transparency. We view these results as interpretable diagnostics of CTRL’s behavior rather than as definitive statements about similarity between groups.


\end{document}